\documentclass[10pt,twocolumn,letterpaper]{article}

\usepackage[pagenumbers]{cvpr} %

\usepackage{times}
\usepackage{epsfig}
\usepackage{graphicx}
\usepackage{amsmath}
\usepackage{amssymb}
\usepackage{booktabs}
\usepackage{multirow}
\usepackage{multicol}
\usepackage{pifont}
\usepackage[toc,page,header]{appendix}
\usepackage{minitoc}
\usepackage[table,x11names,dvipsnames]{xcolor}
\definecolor{LightGrey}{rgb}{0.9,0.9,0.9}
\definecolor{Blue}{rgb}{0.2,0.2,0.9}

\def\prepara{{\vspace{2pt}}}
\newcommand\doubleplus{+\kern-1.3ex+\kern0.8ex}

\newcommand{\cmark}{\ding{51}}%
\newcommand{\xmark}{\ding{55}}%

\usepackage[dvipsnames]{xcolor}

\definecolor{cvprblue}{rgb}{0.21,0.49,0.74}
\usepackage[pagebackref,breaklinks,colorlinks,citecolor=cvprblue]{hyperref}

\def\paperID{304} %
\def\confName{CVPR}
\def\confYear{2024}

\usepackage[dvipsnames]{xcolor}

\title{TIM: A Time Interval Machine for Audio-Visual Action Recognition}

\author{Jacob Chalk$^{1*}$ \quad\quad\quad Jaesung Huh$^{2*}$ \quad \quad\quad Evangelos Kazakos$^{3}$ \\Andrew Zisserman$^{2}$ \quad\quad\quad Dima Damen$^{1}$ \\
$^{1}$University of Bristol \quad
$^{2}$VGG, University of Oxford \quad
$^{3}$ Czech Technical University in Prague \\
}

\begin{document}
\maketitle
\doparttoc %
\faketableofcontents %

\makeatletter{\renewcommand*{\@makefnmark}{}
\footnotetext{$^*$Equal technical contribution.}\makeatother}

\begin{abstract}
Diverse actions give rise to rich audio-visual signals in long videos. 
Recent works showcase that the two modalities of audio and video exhibit different temporal extents of events and distinct labels.
We address the interplay between the two modalities in long videos by explicitly modelling the temporal extents of audio and visual events.
We propose the Time Interval Machine~(TIM) where a modality-specific time interval poses as a query to a transformer encoder that ingests a long video input.
The encoder then attends to the specified interval, as well as the surrounding context in both modalities, in order to recognise the ongoing action.

We test TIM on three long audio-visual video datasets: EPIC-KITCHENS, Perception Test, and AVE, reporting state-of-the-art (SOTA) for recognition.
 On EPIC-KITCHENS, we beat previous SOTA that utilises LLMs and significantly larger pre-training by 2.9\% top-1 action recognition accuracy.
 Additionally, we show that TIM can be adapted for action detection, using dense multi-scale interval queries, outperforming SOTA on EPIC-KITCHENS-100 for most metrics, and showing strong performance on the Perception Test.
Our ablations show the critical role of integrating the two modalities and modelling their time intervals in achieving this performance. Code and models at: \small{{\url{https://github.com/JacobChalk/TIM}}}.
\end{abstract}

\section{Introduction}
\label{sec:intro}

\begin{figure}[t]
    \centering
    \includegraphics[width=\linewidth]{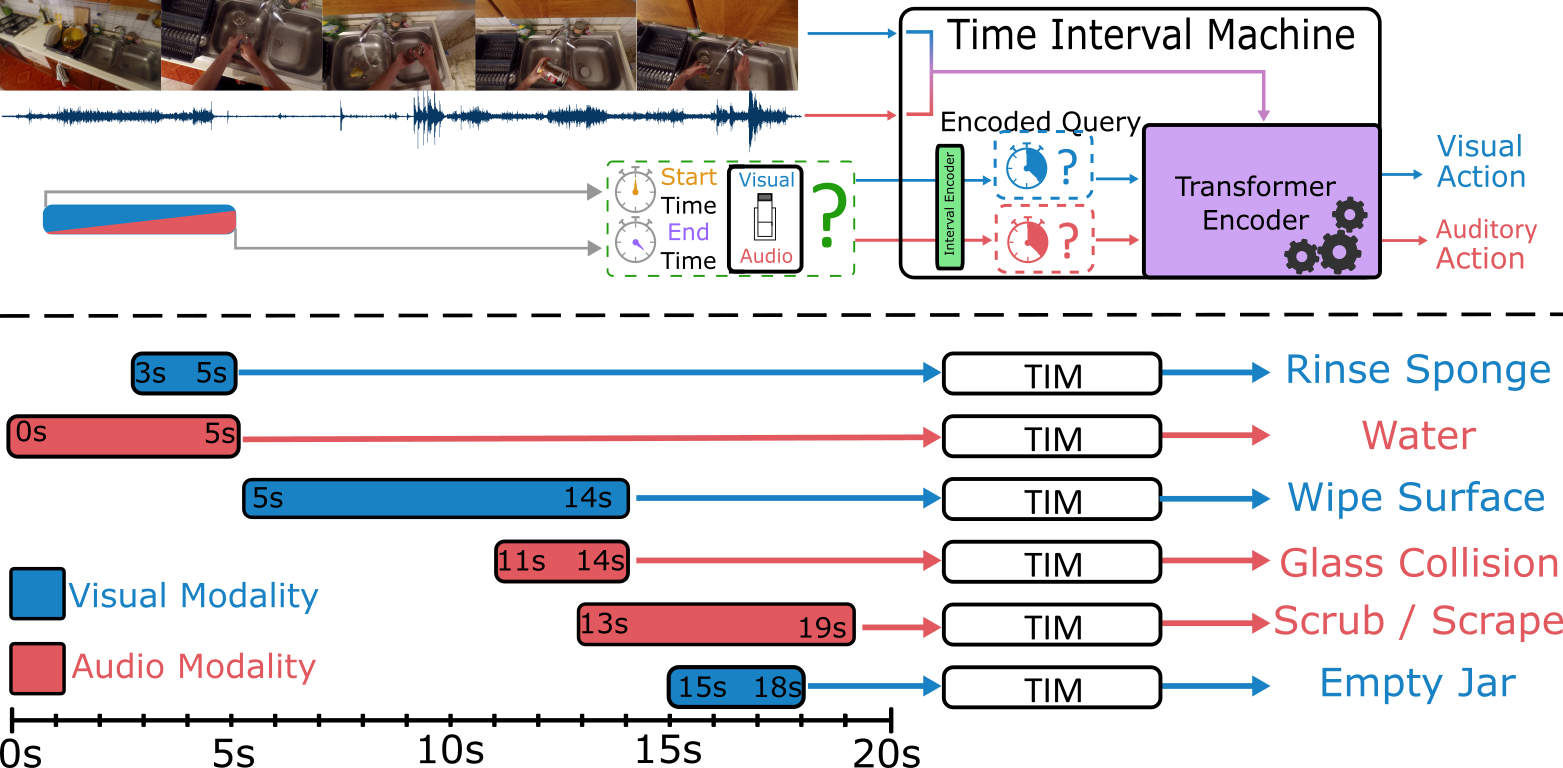}
    \caption{Time Interval Machine~(TIM): \textbf{Top:} Given a visual and auditory stream input, the ongoing action in a particular {\em time interval} is determined by a query specifying the start and end time of the interval, along with the modality of interest.
    \textbf{Bottom:} TIM can query for visual (e.g.\ `Rinse Sponge') and auditory (e.g.\ `Water') action classes, as well as distinguish between overlapping actions within the same modality (`Glass Collision' and `Scrub / Scrape').  } 
    \label{fig:teaser}
\end{figure}

Long videos exhibit a quick succession of auditory and visual events.
The latest attempts to annotate events in these modalities separately~\cite{EPICSOUNDS2023,pătrăucean2023perception}, showcase that both the temporal extents and class labels differ between the two.
However, these events still remain correlated -- identifying temporally close events in both modalities can improve recognition of actions in both visual and audio. 

Furthermore, most methods to date typically only utilise the exact temporal extent of an action;
a precise, trimmed clip of the action is fed into a convolutional~\cite{wang2016temporal, carreira2017quo,Feichtenhofer_2019_ICCV} or transformer-based~\cite{arnab2021vivit,girdhar2022omnivore, liu2022video} backbone, which predicts the action taking place. 
Even when the surrounding context is utilised to improve action recognition~\cite{wu2019long, Sener_2020_ECCV, kazakos2021MTCN},
again, this context is supplied in the form of exact clips of neighbouring actions, rather than the untrimmed long input video. 

In this paper, we propose an approach that encodes multiple events that occur in both visual and auditory streams of a long video input.
We achieve this by elevating {\em time intervals} to first-class citizens, utilising them to specify a query within an accompanying modality.
We term this mechanism a  \textbf{Time Interval Machine~(TIM)}. 
It is able to receive a long video input, and output the actions that occur \textit{within the queried intervals of the queried modalities}.

Consider the example in Figure~\ref{fig:teaser}. The input contains the sound of water running while a sponge is being rinsed, which is then used to wipe a surface. These distinct events may vary significantly in duration and may be more prominent in the audio or visual modality.
Despite differences between these events, there are likely many correlations between them and the surrounding context, which may be beneficial to recognising a given event (e.g.\ the sound of water is relevant to rinsing a sponge, providing useful information for recognising the visual action).
TIM is able to exploit this by accessing the context within both modalities, including the background when no events occur.
It can then distinguish between different, potentially overlapping, events within the same input by querying the time interval of a particular event within a given modality.  

We test TIM on three challenging audio-visual \textbf{recognition} datasets consisting of long videos: EPIC-KITCHENS~\cite{Damen2020RESCALING}, which recently offered distinct audio annotations through EPIC-SOUNDS~\cite{EPICSOUNDS2023}, the Perception Test~\cite{pătrăucean2023perception}, and AVE~\cite{tian2018ave}.
We show that TIM can effectively learn both visual and auditory classes within a long input, outperforming the current SOTA top-1 accuracy on EPIC-KITCHENS by 2.9\% and 1.4\% on EPIC-SOUNDS, despite competing methods for the former using far larger pre-training datasets, large language models or higher resolution input.
We also outperform models pre-trained with public datasets on AVE by 0.6\% and improve over a strong baseline on the Perception Test in visual and audio action recognition by 9.9\% and 3.2\% respectively.

Additionally, we adapt TIM to action \textbf{detection}, through fixed multi-scale dense querying with an added \textit{interval regression} loss. We report strong detection results on EPIC-KITCHENS and the Perception Test, outperforming Action Former~\cite{zhang2022actionformer} by 1.6 and 4.3 mAP respectively. 

Our contributions are summarised as:
(i) we propose the TIM query mechanism for attending to modality-specific intervals in long videos.
(ii) we efficiently train TIM to encode/query multiple audio-visual actions using time intervals.
(iii) we showcase the value of TIM for both visual and auditory action recognition, and adapt it for detection with an added interval regression loss.
(iv) we achieve new SOTA in both video and multi-modal recognition on multiple datasets.

\section{Related Works}
\label{sec:related_works}
\noindent\textbf{Audio-visual action recognition}. A number of works have employed audio and visual modalities for action recognition~\cite{xiao2020audiovisual,Gao_2020_CVPR, Nagrani21c,Wang_2020_CVPR,kazakos2019TBN,kazakos2021MTCN}. 
Some introduce new architectures to effectively fuse modalities~\cite{kazakos2019TBN, xiao2020audiovisual,Nagrani21c,kazakos2021MTCN};
others propose unique training techniques to solve problems occurring while training multi-modal models, such as Gradient Blending~\cite{Wang_2020_CVPR}, to tackle overfitting at different speeds for each modality, or contrastive learning for cross-modal discrimination~\cite{morgado2021audio}.
However, these works use the same set of semantic and temporal labels for both modalities. 
Recent works have shown that both the temporal intervals and semantics of events differ between modalities~\cite{EPICSOUNDS2023,pătrăucean2023perception}. 
\cite{tian2020avvp} temporally annotates visual and auditory events independently, although they share the same set of labels.
In this work, we train with distinct labels for each modality to leverage discriminative audio and visual actions.

\noindent\textbf{Leveraging temporal context}. Several works have considered incorporating temporal context \cite{Ng_2019_arXiv,zhang_2021_CVPR,wu2019long,wu2022memvit,kazakos2021MTCN}, a direction orthogonal to employing multiple modalities and particularly useful in untrimmed videos. 
An auto-regressive LSTM-based encoder-decoder is proposed in \cite{Ng_2019_arXiv} for action sequence classification, effectively leveraging past action context to predict the current action. 
The Temporal Query Network~\cite{zhang_2021_CVPR} uses learnable query vectors that correspond to specific attributes of a long video, allowing the model to attend to the aspects of the video and its surrounding context to produce a response for each attribute.  \cite{wu2019long}~proposes to enhance the representation of the action by aggregating temporal context from neighbouring action clips using a Long-Term Feature Bank along with an attention mechanism. 
\cite{wu2022memvit} crafts a more sophisticated memory bank by storing keys and values of all the intermediate layers of a transformer to aggregate the past context. 
Lastly, \cite{kazakos2021MTCN} exploits multi-modal temporal context from surrounding actions using vision, audio, and language. 

\cite{wu2022memvit,wu2019long,kazakos2021MTCN} are the closest to our approach, in that the common goal is to enrich the representation of the action of interest using surrounding context from the untrimmed video, rather than neighbouring clips. Nevertheless, \cite{wu2022memvit,wu2019long} are single modality models, recognising visual actions solely. \cite{kazakos2021MTCN} assumes the temporal extents of all actions are known, including for the test set, which is restrictive.

\noindent\textbf{Queries in visual models}. 
Learning visual queries with Transformer architectures has gained recent attention~\cite{carion_2020_ECCV,Locatello_2020_NeurIPS,zhang_2021_CVPR,herzig2022promptonomyvit,jia2022vpt}. Commonly, approaches employ a set of learnable vectors that are used to inquire about the presence of a concept in the input. For example, in \cite{Locatello_2020_NeurIPS,carion_2020_ECCV} the learnable queries correspond to different objects, whereas in \cite{herzig2022promptonomyvit} they are used for multi-task learning and each learnable query corresponds to a different task. \cite{jia2022vpt} has incorporated learnable queries for adapting a pre-trained model while keeping the rest of its parameters frozen. Closest to our motivation is \cite{zhang_2021_CVPR}, where the queries correspond to events and their attributes for fine-grained action recognition in videos. The authors note that the queries also have the role of temporally localising the events in untrimmed videos. 

Different from~\cite{zhang_2021_CVPR} and other works, our queries are primarily temporal with no semantic interpretation and 
are applied to multiple modalities.
Importantly, since time is continuous, we cannot use a predefined set of queries.
Instead, we employ an MLP architecture to encode time, in a form akin to a universal clock. 
We present our approach next.

  \begin{figure*}[t]
  \centering
  \includegraphics[width=\linewidth]{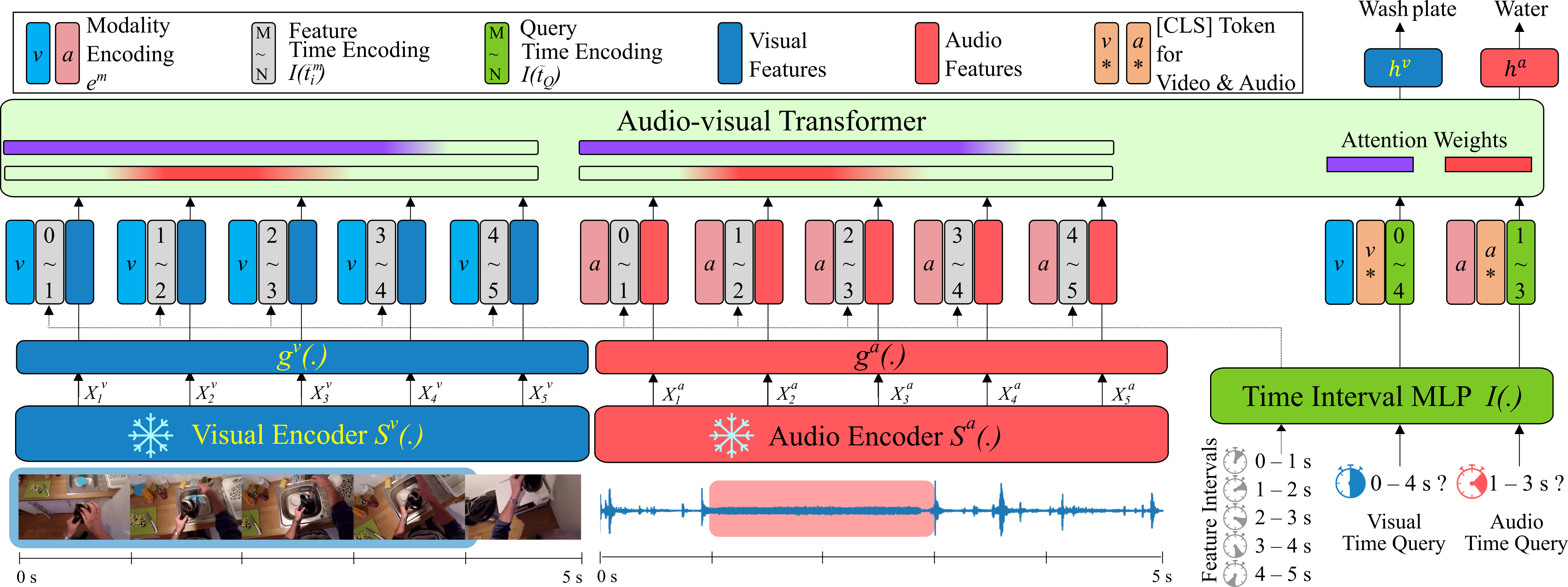}
  \caption{\textbf{Overview of the Time Interval Machine~(TIM)}. The model ingests a sequence of audio and visual features from a video, with each feature time-stamped by the temporal interval it spans, and encoded with its modality. To infer the action occurring over a temporal interval (a visual or audio event) a query is formed specifying the interval and modality of interest.}
  \vspace*{-12pt}
  \label{fig:model_arch}
\end{figure*}

\section{Time Interval Machine}
\label{sec:time-interval-machine}
 In this section, we describe the \textbf{Time Interval Machine} (TIM), a multi-modal transformer encoder architecture where all inputs, both features and queries, are encoded with their associated {\em time intervals}. A time interval incorporates the duration and position of each audio and visual feature and is also used to {\em query} the network for any action occurring within the given time interval. 
 
 The architecture of TIM is illustrated in Figure~\ref{fig:model_arch}. 
 It ingests a large video input, represented as a sequence of audio and visual features, and outputs the ongoing auditory or visual action label for the provided query time intervals.

 \subsection{Model architecture}
 \label{subsec:audio-visual transformer}
 \noindent \textbf{Input.} 
The input to TIM is a long crop of the untrimmed video, represented by extracted features.
When considering two modality inputs, such as video and audio, each modality is embedded separately as follows:
for each modality $m$, let $\mathbf{X}^{m} = [X^{m}_{1},\cdots, X^{m}_{N^{m}}]$ be $N^{m}$ temporally-ordered feature representations of the input video, obtained from a pre-trained feature extractor $S^{m}(\cdot)$. 
We feed the features through modality-specific embedding layers $g^{m}(\cdot)$, projecting them to a lower, common dimension $D$ across all modalities. 
The embedded features\footnote{Note that the number of features can differ between modalities} are then tagged with modality encodings and time interval encodings, forming the input to the transformer encoder. 
We now detail how we encode the time intervals.

\begin{figure}[t]
  \centering
  \includegraphics[width=0.9\linewidth]{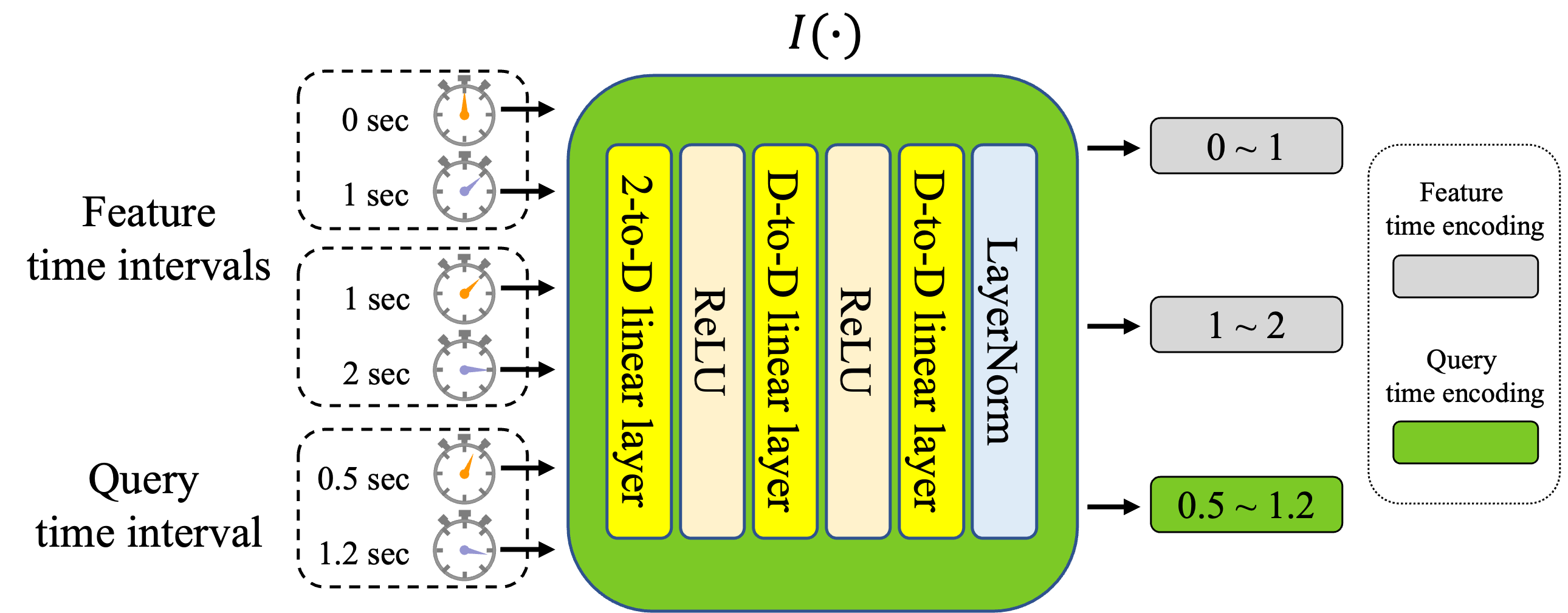}
  \caption{Illustration of the Time Interval MLP $I(\cdot)$. It inputs the \textbf{two} dimensional vector, start and end times of an interval, and produces a single vector, which can be concatenated along the channel dimension to either input features or \texttt{[CLS]} tokens. The figure shows three time interval inputs and three corresponding outputs. Note that in practice, time intervals are ingested simultaneously.}
  \vspace*{-12pt}
  \label{fig:Time Interval_mlp}
\end{figure}

\prepara\noindent\textbf{Encoding Time Intervals.}
In this work, we introduce a new type of learnt query network, the \textbf{Time Interval MLP}, which produces a single $D$-dimensional vector representing a given time interval.
This network is used within TIM to encode the time intervals of the input features and the time interval we wish to query, and later classify.
Figure~\ref{fig:Time Interval_mlp} illustrates the concept of this network.

The Time Interval MLP $I(\cdot) : \mathbb{R}^{2} \rightarrow \mathbb{R}^{D}$ receives a time interval, represented by start and end time, and produces a single $D$-dimensional encoding. 
Note that this is distinct from encoding the start and end times separately. 
Specifically, let $t_{s}$ and $t_{e}$ be the start and end time of an interval of interest, normalised by the length of the long video input.
$I(\cdot)$ receives the interval $\tilde{t} = [t_s,t_e]$ as input, and outputs a $D$-dimensional vector encoding of that interval.
This vector encodes both the relative position of the time interval within the input, as well as its duration. 
This vector then acts as a \textit{query} for the model concerning the action taking place within the interval. Furthermore, each feature $\{ X^{m}_{i} \}$ spans a certain time interval within the input.
Thus, it is important to also encode the time intervals of the features.

In summary, the Time Interval MLP acts as an \textit{universal clock}, which encodes the temporal extent of features, from any modality, within the input.
Note that it is critical that the same Time Interval MLP is used for encoding all time intervals of the input features and queries across both modalities to accurately encode universal time. 
It is also important to note that Time interval MLP can cover \textit{continuous} time intervals, whereas traditional positional encoding only covers a fixed set of positions of the input features.
The Time Interval MLP is trained end-to-end along with the transformer.

\prepara\noindent\textbf{Transformer Feature Inputs.}
Let $\mathbf{\tilde{t}}^{m} = [\tilde{t}^{m}_{1},\cdots,\tilde{t}^{m}_{N^{m}}]$ be the corresponding time intervals of the video's features $\mathbf{X}^{m}$ from the modality $m$.
We inject the encoded time interval $I(\tilde{\mathbf{t}}^{m})$ into the embedded features via channel-wise concatenation.
A learnable modality-specific encoding $e^{m}\in \mathbb{R}^{2D}$ is then summed to the temporally-encoded features to discriminate between each modality. 
In summary, the feature inputs $\mathbf{E}^{m}$ for TIM are computed by,
\begin{align}
  E^{m}_{i} = [g^{m}(X^{m}_{i}), I(\tilde{t}^{m}_{i})] + e^{m} \quad \forall i\in[1,...,N^{m}]
  \label{eq:av_encoder}
\end{align}
where $[\cdot,\cdot]$ indicates concatenation.

\prepara\noindent\textbf{Transformer Query Inputs.}
To query for an action within an interval of interest, we adopt a standard approach of appending a learnable classification token to the input sequence, $\texttt{CLS}^{m}$. 
 If $\tilde{t}_{Q}$ is an interval of interest, we concatenate the time interval representation $I(\tilde{t}_{Q})$ to this classification token along the channel dimension, which acts as a query for the network in order to predict the corresponding action happening within $\tilde{t}_{Q}$.
 We also add the modality-specific encoding $e^{m}$ to each classification token, as a flag to distinguish between which modality we are querying. 
 The encoded $\texttt{[CLS]}^{m}$ tokens can be more formally defined as:
\begin{equation}
  \texttt{[CLS]}^{m} = [\texttt{CLS}^{m}, I(\tilde{t}_{Q})] + e^{m} 
  \label{eq:cls_tokens}
\end{equation}
During training, we add a classification token for \emph{each} action within the input video, resulting in multiple $\texttt{[CLS]}$ tokens across both modalities.

\prepara\noindent\textbf{Transformer Encoder.}
We use a transformer encoder to perform self-attention on the input sequence to aggregate relevant temporal context and cross-modal relations.

 We form the transformer input sequence with the encoded feature inputs $\textbf{E}^{m}$ and \textbf{one or more} classification tokens $\texttt{[CLS]}^{m}$, representing each time interval query, and feed these into the encoder. 
 Note that we recognise all actions from any modality simultaneously by appending multiple $\texttt{CLS}^{m}$ tokens to the input.
The transformer output representation of $\texttt{[CLS]}^{m}$, namely $Z^{m}_{\texttt{CLS}}$, is then passed to the corresponding linear classifier to predict the action labels. 

Importantly, we use an attention mask to prevent queries from attending to one another, and we similarly prevent input features from attending to queries.
This ensures each query is recognised without the privileged knowledge of any other query, or action boundary, during inference.

 \subsection{Training and Testing in TIM}
 \label{subsec:loss}
 To train TIM, we consider all long segments of $W$ seconds and stride $H_{w}$ across the entirety of the untrimmed videos.
 We randomly select batches from these.
For each window, we query all annotated audio and visual actions that overlap with the window by more than $\delta=0.2$ seconds.

All queries in the window are encoded and concatenated to separate $\texttt{CLS}$ tokens. To classify queries, let $h^{m}_{\texttt{CLS}}(\cdot)$ be a linear classifier for modality $m$, and let $\hat{y}^{m}_{\texttt{CLS}} = h^{m}(Z^{m}_{\texttt{CLS}})$ be the predicted action of the output representation $Z^{m}_{\texttt{CLS}}$. 
We train TIM by using a cross-entropy classification loss $CE(\cdot)$ on the ground truth $y^{m}_{\texttt{CLS}}$ by:
 \begin{equation}
  L^{m} = \frac{1}{N_{Q}} \sum^{N_{Q}} CE (\hat{y}^{m}_{\texttt{CLS}}, y^{m}_{\texttt{CLS}})
  \label{eq:modality_loss}
\end{equation}
where $N_{Q}$ is the number of queries within the batch.

\prepara\noindent\textbf{Temporal Distance Loss.}
 In addition to the standard classification loss, we introduce a Temporal Distance (TD) loss as an auxiliary loss for training TIM. 
Inspired by~\cite{liu2021efficient}, where relative patch positions in token embeddings are learnt using self-supervision, we similarly train the network to take two transformer outputs and predict the elapsed time between their corresponding time intervals.

Let the $\mathbf{Z}_{1:\sum_{m}{N^{m}}}$ be the transformer outputs of the features from all modalities. 
We randomly sample a set of feature pairs $\mathbb{B} \subset \mathbf{Z}_{1:\sum_{m}{N^{m}}}$ from these outputs, concatenate along the channel dimension and feed them to the temporal distance regression head $h_{\tilde{t}} (\cdot) : \mathbb{R}^{4D} \rightarrow \mathbb{R}^{1}$ to predict the time interval difference between each pair.
Note that feature pairs can be sampled both \textit{within} and \textit{across} modalities. In our case, we sample across modalities by pairing one visual feature with another audio feature. 
This helps the model to learn the temporal relations between modalities.

Formally, a TD loss $L^{td}$ is computed as:
\begin{equation}
  L^{td}= \sum_{\{Z_{i}, Z_{j}\} \in \mathbb{B}}{\left|h_{\tilde{t}}(Z_{i},  Z_{j}) - d_{ij}\right|}
  \label{eq:aux_loss}
\end{equation}
where $d_{ij}$ is the temporal distance between intervals $\tilde{t}_i, \tilde{t}_j$.

 \prepara\noindent\textbf{Training objective and regime.}
 For our final training loss, we sum the losses across modalities along with the TD loss:
 \begin{align}
  L^{total} = \left(\sum_{m \in \mathbb{M}}{\lambda^{m} L^{m}}\right) + \lambda^{td} L^{td}
  \label{eq:total_loss}
\end{align}
 where $\mathbb{M}$ is a set of modalities, $\lambda^{m}$ controls the strength of each modality's loss and $\lambda^{td}$ is a hyperparameter that controls the strength of the TD loss.

\prepara\noindent\textbf{Test-Time Augmentation.}
We use test-time augmentation, as this generally increases prediction robustness and performance~\cite{shanmugam2021better, patrick2021keeping}.
In TIM, we use a sliding window across the untrimmed video, thus feeding the same interval query with varying contexts. 
We then aggregate predictions of the same interval query across windows to make the final prediction.

\subsection{Adapting for Detection}
While primarily designed for recognition, we can adapt TIM for detection. The backbone remains largely unchanged from recognition, but there are two main differences. First, we construct dense multi-scale interval queries spanning the entirety of the video input at each scale. These are used as interval queries in both training and detection inference. The multi-scale intervals allow for detecting both long and short actions. Second, we introduce an additional interval regression head, which regresses the query interval to the action's exact temporal duration. 

During training, we deem any query in the multi-scale pyramid that overlaps with a ground truth action by more than some IoU threshold as a positive query.
In addition to classifying the query, we train a DIOU regression loss~\cite{zheng2020diou} to predict the exact interval of the action.
Both classification and interval regression losses are trained jointly. 
We provide full details in the ArXiv appendix.

\section{Experiments}
\label{sec:experiments}

This section describes the datasets used to evaluate our model, implementation details and results along with a comparison to the state-of-the-art methods. 

\subsection{Dataset}
\label{subsec:dataset}

\prepara\noindent\textbf{EPIC-KITCHENS-100~\cite{Damen2020RESCALING}}
is a large-scale video dataset, including 700 egocentric videos recording actions in kitchens.
It consists of 89,977 segments of fine-grained actions. %
Inspired by prior works~\cite{girdhar2022omnivore, sudhakaran2021saic_cambridge, tai2022nvidia}, we directly predict the action out of 3806 classes present in the train and validation set to avoid predicting invalid actions.

\prepara\noindent\textbf{EPIC-SOUNDS~\cite{EPICSOUNDS2023}}
 offers audio annotations capturing temporal extents and class labels within the audio stream of EPIC-KITCHENS-100.
The annotations contain 78,366 labelled audio events.
We combine the visual annotations from EPIC-KITCHENS with the audio annotations from EPIC-SOUNDS to train our audio-visual model.
TIM can recognise actions from both datasets using a single model.

\prepara\noindent\textbf{AVE~\cite{tian2018ave}} contains 4,143 videos covering a range of real-life scenes and labelled with 27 categories, such as church bell, male speaking and dog barking. 
Each video is equally divided into 10 segments, each 1 second in length.
We evaluate TIM on the supervised audio-visual event localisation task.
Given a 1 second segment, we recognise the ongoing action out of the 27 categories \textit{plus} a background class.

\prepara\noindent\textbf{Perception Test~\cite{pătrăucean2023perception}} is a recent multimodal video benchmark of 11,620 videos with an average length of  23 seconds, and provides both temporal action and sound annotations. There are 73,503 visual annotations spanning 63 classes, versus 137,128 sound annotations over 16 classes.

\subsection{Implementation Details}
\label{subsec:implementation_details}

\prepara\noindent\textbf{Architectural Details.}
Visual and audio embedding layers $g_m$ consist of a single 512-D feed-forward layer, followed by a GELU~\cite{hendrycks2016gaussian} activation and layer normalisation~\cite{ba2016layer} are used to project the features to a common space.
The Time Interval MLP $I$ consists of three linear layers with 512-D hidden dimension, followed by ReLU activations, with layer normalisation after the output of the last linear layer.
We include 512-D  learnable \texttt{[CLS]} tokens: $\texttt{[CLS]}_{action}^{m}$ for each query in each modality which become 1024-D after concatenation with the encoded time interval. They are then summed with 1024-D modality encodings; $e^{m}$. 

The audio-visual transformer contains four encoder layers, each with 8 attention heads, GELU activations, and 1024-D keys, queries and values.
A dropout rate of $p=0.1$ is applied within the encoder layers.
We also apply channel-wise dropout with $p=0.5$ directly to the raw input features, as well as to the encoded transformer input.
The temporal distance head consists of two linear layers with a hidden dimension of 1024 and a third which outputs a single number corresponding to the elapsed time between each time interval. We include the architectural ablations on encoder layers and the temporal distance head in the ArXiv appendix.

\prepara\noindent\textbf{Training / Validation Details.}
We train each model for 100 epochs, usingAdamW~\cite{loshchilov2017decoupled} with a batch size of 64 and a weight decay of 1e-4.
A linear learning rate warm-up is applied for the first two epochs, starting from 1e-6 to a target learning rate, and we use a cosine learning rate scheduler.
We set the TD loss weight $\lambda^{td}$ to 0.3.
We pad the queries for each window in the batch to the maximum number of queries in a single window in each dataset.
We provide implementation details per dataset in the ArXiv appendix.

\subsection{Results}
\label{subsec:results}
We compare TIM with SOTA models for each dataset. %

\prepara\noindent\textbf{EPIC-KITCHENS / EPIC-SOUNDS Results.}
We train a single model on both visual and audio labels of the EPIC-KITCHENS videos, reporting results on both datasets.

For the visual features, we concatenate Omnivore~\cite{girdhar2022omnivore} and VideoMAE-L~\cite{tong2022videomae} features
along the channel dimension, forming 2048-D features.
For the audio features, we use Auditory SlowFast~\cite{kazakos2021slow}, which generalises well across diverse audio domains~\cite{wang2022towards}. 
For both modalities, we extract 1 second features every 0.2s.
For training, we extract additional augmented feature sets - with RandAugment~\cite{cubuk2020randaugment} for visual and SpecAugment~\cite{park2019specaugment} for audio features. 

Table~\ref{tab:sota_epic100} compares TIM with the SOTA models on EPIC-KITCHENS-100.
We outperform M\&M Mix~\cite{xiong2022m} by 5.1\% on verb, 0.9\% on noun and 3.9\% on action.
Compared to our model, both MTV and M\&M Mix are trained with an additional private dataset~\cite{stroud2020learning} which contains 194K hours of 70 million videos while we only use the open-source visual backbone pre-trained with public datasets.
We also outperform LaViLa~\cite{zhao2023lavila} and AVION~\cite{zhao2023training} which leverage pre-trained LLMs to learn video representations.

We note that we outperform all prior works, often without additional techniques that boost performance.
For example, we use short-sided cropped 224$\times$224 images while ~\cite{xiong2022m} uses 420$\times$420, which enlarges the spatial resolution of objects in the egocentric video, enabling better noun recognition.
We expect a further performance boost when implementing any of: higher resolution feature extractors, additional large-scale pre-training and the introduction of a LLM.
We leave this as an avenue for future work.

Table~\ref{tab:sota_epicsounds} compares TIM against prior results on EPIC-SOUNDS, where  
TIM outperforms SOTA by 1.4\%. 

For detection, we show that TIM can produce competitive results when compared to models primarily designed for this task in Table~\ref{tab:sota_epic100_det}. TIM adapted for detection outperforms ActionFormer~\cite{zhang2022actionformer} by 2.3 mAP on verb and 1.6 mAP on noun using the same set of features.

\begin{table}[t]
\centering
\resizebox{\linewidth}{!}{
\begin{tabular}{l c c c c c }
\hline
Model & $xp$ & LLM & Verb & Noun & Action \\
\hline
\rowcolor{LightGrey} \multicolumn{6}{l}{\textit{Visual-only models}} \\	
MFormer-HR~\cite{patrick2021keeping} & 336p & \xmark & 67.0 & 58.5 & 44.5 \\
MoViNet-A6~\cite{kondratyuk2021movinets} & 320p & \xmark & 72.2 & 57.3 & 47.7 \\
MeMViT~\cite{wu2022memvit} & 224p & \xmark & 71.4 & 60.3 & 48.4 \\
Omnivore~\cite{girdhar2022omnivore} & 224p & \xmark & 69.5 & 61.7 & 49.9  \\
MTV~\cite{yan2022multiview}& 280p & \xmark & 69.9 & 63.9 & 50.5 \\
LaViLa (TSF-L)~\cite{zhao2023lavila} & 224p & \cmark & 72.0 & 62.9 & 51.0 \\
AVION (ViT-L)~\cite{zhao2023training} & 224p & \cmark & 73.0 & 65.4 & 54.4 \\
\textbf{TIM (ours)} & 224p & \xmark & \textbf{76.2} & \textbf{66.4} & \textbf{56.4} \\
\hline 
\rowcolor{LightGrey} \multicolumn{6}{l}{\textit{Audio-visual models}}\\
TBN~\cite{kazakos2019TBN} & 224p & \xmark & 66.0 & 47.2 & 36.7 \\ 
MBT~\cite{Nagrani21c} & 224p & \xmark & 64.8 & 58.0 & 43.4 \\
MTCN~\cite{kazakos2021MTCN} & 336p & \xmark & 70.7 & 62.1 & 49.6 \\
M\&M~\cite{xiong2022m} & 420p & \xmark & 72.0 & 66.3 & 53.6 \\
\textbf{TIM (ours)} & 224p & \xmark & \textbf{77.1} & \textbf{67.2} & \textbf{57.5} \\
\hline 
\end{tabular}
}
\caption{Comparisons to state-of-the-art {\em recognition} models on the EPIC-KITCHENS validation set. We report the top-1 accuracy for verb, noun and action (\%). LLM: large language model is used during pre-training. $x$p: input resolution of $x \times x$.}
\label{tab:sota_epic100}
\end{table}

\begin{table}[!t]
\centering
\resizebox{\linewidth}{!}{
\begin{tabular}{c c c c c c}
\hline
\textbf{Model} & SSAST~\cite{gong2022ssast} &  ASF~\cite{kazakos2021slow} & DiffSED~\cite{bhosale2023diffsed} &\textbf{TIM (A)} &\textbf{TIM (A+V)} \\
\hline 
\textbf{Top-1 acc} & 53.5 & 53.8 & 56.9 & 55.7 & \textbf{58.3}\\
\hline 
\end{tabular}
}
\caption{Comparisons to state-of-the-art {\em sound recognition} models on EPIC-SOUNDS. We report the top-1 accuracy (\%) on Val. The performance of SSAST and ASF are from ~\cite{EPICSOUNDS2023}.}
\label{tab:sota_epicsounds}
\end{table}

\begin{table}[t]
    \centering
    \resizebox{\linewidth}{!}{
        \begin{tabular}{| c c c| c | c c c c c | c|}
            \hline
             \multirow{2}{*}{\textbf{Model}} & \multirow{2}{*}{V} & \multirow{2}{*}{A} & \multicolumn{7}{c|}{\textbf{Average Precision (AP)}} \\
             \cline{4-10} & & & Task & @0.1 & @0.2 & @0.3 & @0.4 & @0.5 & Avg. \\\hline \multirow{2}{*}{G-TAD~\cite{xu2020gtad}} & \multirow{2}{*}{\checkmark} & \multirow{2}{*}{\xmark} & Verb & 12.1 & 11.0 & 9.4 & 8.1 & 6.5 & 9.4  \\
             & & & Noun & 11.0 & 10.0 & 8.6 & 7.0 & 5.4 & 8.4 \\
             \hline 
             \multirow{2}{*}{ActionFormer~\cite{zhang2022actionformer}} & \multirow{2}{*}{\checkmark} & \multirow{2}{*}{\xmark} & Verb & 26.6 & 25.4 & 24.2 & 22.3 & 19.1 & 23.5  \\
             & & & Noun & 25.2 & 24.1 & 22.7 & 20.5 & 17.0 & 21.9 \\
            \hline 
             \multirow{2}{*}{ActionFormer - Our Features} & \multirow{2}{*}{\checkmark} & \multirow{2}{*}{\xmark} & Verb & 29.6 & 28.8 & 26.9 & 24.4 & 21.6 & 26.3   \\
             & & & Noun & 34.3 & 32.6 & 30.2 & 27.4 & 22.6 & 29.4 \\
            \hline 
            \multirow{2}{*}{TIM} & \multirow{2}{*}{\checkmark} & \multirow{2}{*}{\checkmark} & Verb & \textbf{32.9} & \textbf{31.6} & \textbf{29.6} & \textbf{27.0} & \textbf{22.2} & \textbf{28.6} \\
            & & & Noun & \textbf{36.4} & \textbf{34.8} & \textbf{32.1} & \textbf{28.7} & \textbf{22.7} & \textbf{31.0} \\
            \hline
        \end{tabular}
    }
    \caption{Comparisons to state-of-the-art {\em detection} models on the EPIC-KITCHENS validation set. We report the average precision at IOU thresholds $[0.1, 0.2, 0.3, 0.4, 0.5]$ as well as their average across all thresholds on verb, noun.}
    \vspace*{-12pt}
    \label{tab:sota_epic100_det}
\end{table}

\prepara\noindent\textbf{AVE Results.} 
As this dataset contains joint audio-visual labels, we train TIM 
by duplicating the query, i.e.\ using a $\texttt{[CLS]}$ for each modality, and combine their logits during training and inference. We use the pre-trained publicly available models from~\cite{tian2018ave} for a fair comparison with other works.
We also apply AVGA~\cite{tian2018ave} to spatial visual features from VGG-19 before feeding them to the transformer.

Table~\ref{tab:sota_ave} shows our results on the AVE dataset.
Combining audio and video significantly improves the performance on TIM.
The results from~\cite{feng2023css} perform best but could not be replicated. 
We also report TIM using the Omnivore visual features and Auditory Slowfast features used for EPIC-KITCHENS, which achieves a 0.6\% boost in performance.

\begin{figure*}[h!]
  \centering
  \includegraphics[width=\linewidth]{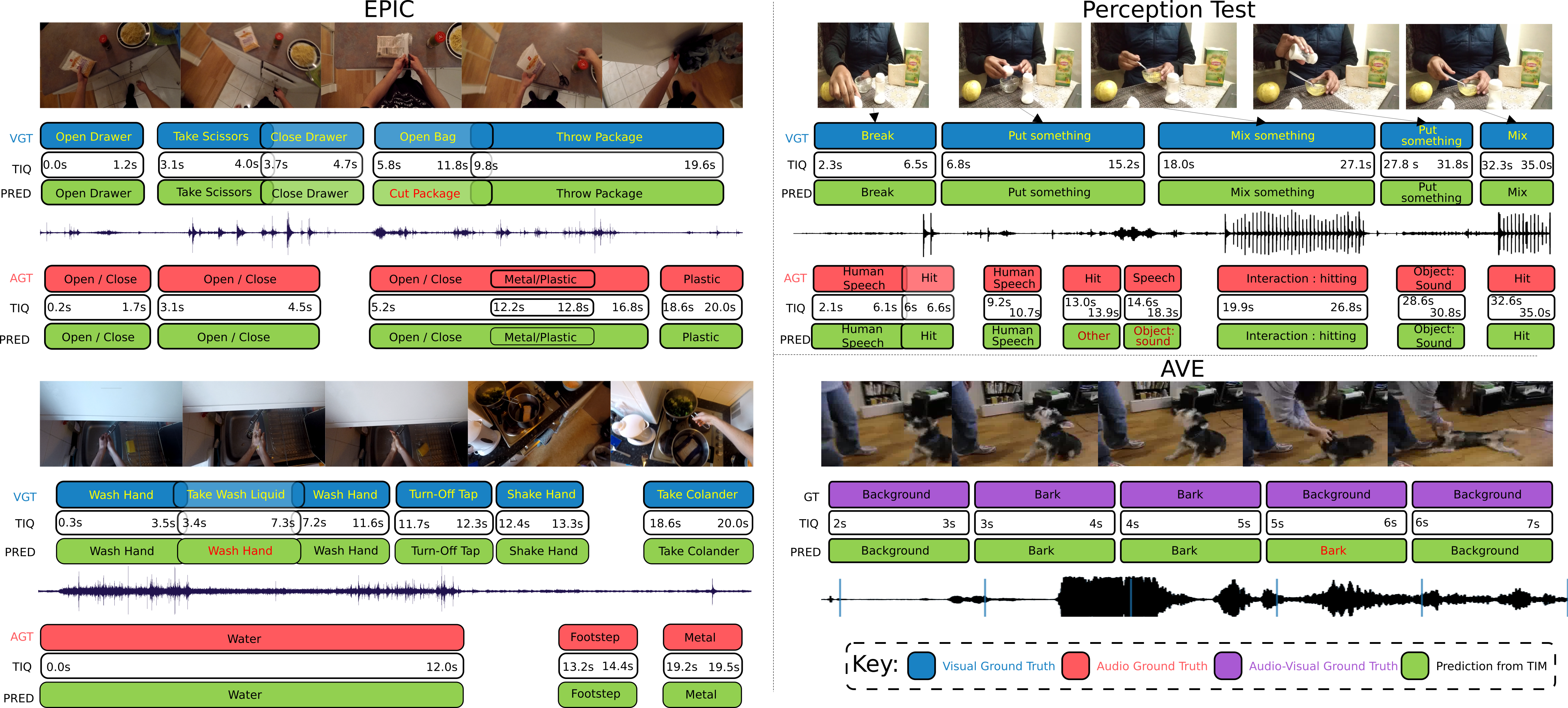}
  \caption{Qualitative results for all datasets. \textbf{PRED:} Prediction by TIM, \textbf{TIQ:} Time Interval Queries, \textbf{V/AGT:} Visual/Audio Ground Truth.}
  \label{fig:qual_results}
  \vspace{-10pt}
\end{figure*}

\begin{table}[!t]
\centering
\resizebox{\linewidth}{!}{
\begin{tabular}{cccccccc}
\hline
\textbf{Model} & PSP & CPSP & CSSNet
&\multicolumn{4}{c}{TIM}\\ \cline{5-8}
&\cite{zhou2021PSP} &\cite{zhou2022CPSP} &\cite{feng2023css}$\dagger$ &V &A &AV &AV$\star$\\
\hline 
\textbf{Top-1 acc} & 77.8 & 78.6 & \textbf{80.5} & 62.8 & 65.5 & 79.2 &79.8\\
\hline 
\end{tabular}}
\caption{Top-1 event classification accuracy (\%) on AVE Test set. $\dagger$: no official code or public model provided to replicate results. We show the models trained only with publicly available datasets. $\star$: results with Omni+ASF features.}
\label{tab:sota_ave}  
\end{table}

\prepara\noindent\textbf{Perception Test Results.}
We use the same backbone for the Omnivore features and Auditory Slowfast features and train a single model using both the visual and audio labels.
Table~\ref{tab:perception_recognition} compares the results on the newly introduced Perception Test.
We train an MLP classifier, with two linear layers and a ReLU activation, on the features directly as a baseline. 
We also evaluate on an audio-visual model that does use context with MTCN. Compared to these methods TIM clearly shows significant improvements. 
Results are improved over MTCN by 9.9\% and 3.2\% on visual and audio recognition tasks respectively.
We also provide detection results in Table~\ref{tab:sota_perception_det}. TIM improves over ActionFormer~\cite{zhang2022actionformer} by 3.3 average mAP on visual actions and by 0.9 average mAP on sound, when using the same features.

\begin{table}[t]
 \centering
    \resizebox{\linewidth}{!}{
        \begin{tabular}{c c c c c}
        \hline
        \rowcolor{LightGrey} \multicolumn{5}{l}{\textit{Perception Test Action}} \\ 
        \textbf{Model} & MLP (V) & MTCN~\cite{kazakos2021MTCN}(A+V) & \textbf{TIM (V)} & \textbf{TIM (A+V)} \\
        \hline 
        \textbf{Top-1 acc} & 43.7 & 51.2 & 56.1 & \textbf{61.1}\\
        \hline
        \rowcolor{LightGrey} \multicolumn{5}{l}{\textit{Perception Test Sound}} \\ 
        \textbf{Model} & MLP (A) & MTCN~\cite{kazakos2021MTCN}(A+V) & \textbf{TIM (A)}  & \textbf{TIM (A+V)} \\
        \hline 
         \textbf{Top-1 acc} & 50.6 & 52.9 & 54.8 & \textbf{56.1} \\
         \hline 
        \end{tabular}
    }
\caption{Comparisons to trained recognition baselines on the Perception Test validation split. We show both action and sound recognition and the benefit of including audio-visual in TIM for both challenges. \textbf{V} : visual and \textbf{A} : audio input features. MLP is the result by training an MLP classifier with the features directly.}
\label{tab:perception_recognition}
\end{table}

\begin{table}[t]
    \centering
    \resizebox{\linewidth}{!}{
    \centering
    \begin{tabular}{|c|c c c c c | c |}
    \hline
     \multirow{2}{*}{\textbf{Model}} & \multicolumn{6}{c|}{\textbf{Average Precision (AP)}} \\
     \cline{2-7} 
     & @0.1 & @0.2 & @0.3 & @0.4 & @0.5 & Avg. \\
     \hline 
    \rowcolor{LightGrey} \textit{Perception Test Action} & \multicolumn{5}{c|}{} & \\  
     ActionFormer~\cite{zhang2022actionformer} & 27.8 & 27.6 & 25.2 & 23.0 & 20.0 & 24.5 \\
     TIM & \textbf{33.5} & \textbf{32.2} & \textbf{29.8} & \textbf{26.4} & \textbf{22.0} & \textbf{28.8} \\ 
     \hline
    \rowcolor{LightGrey} \textit{Perception Test Sound} & \multicolumn{5}{c|}{} & \\       ActionFormer~\cite{zhang2022actionformer} & 34.7 & 31.3 & 27.5 & 22.7 & \textbf{17.7} & 26.8\\
     TIM & \textbf{37.5} & \textbf{33.1} & \textbf{27.9} & \textbf{22.8} & 17.2 & \textbf{27.7} \\
     \hline
    \end{tabular}
    }
\caption{Comparisons to strong {\em detection} models on the Perception Test validation set for action and sound localisation. We report the average precision at IOU thresholds $[0.1, 0.2, 0.3, 0.4, 0.5]$ as well as the average across all thresholds.} 
\vspace*{-12pt}
\label{tab:sota_perception_det}
\end{table}

\prepara\noindent\textbf{Cross-Modality in TIM.} When referring to our previous results, we see that including the additional modality provides a performance boost in all cases, highlighting TIM's ability to utilise 
and distinguish between different modalities. 
For example, on EPIC-KITCHENS-100, including audio improves visual action accuracy by 0.9\%.
For EPIC-SOUNDS, the visual modality further improves accuracy by 2.6\%. In the Perception Test, including the audio modality improves visual recognition by 5.0\%, and visual increases sound recognition by 1.3\%. Finally, for AVE, we see a significant improvement, where an audio-visual model increases accuracy by 13.7\% from audio-only.

\prepara\noindent\textbf{Qualitative Results.} We present qualitative results in Figure~\ref{fig:qual_results}. We see that in EPIC-KITCHENS, TIM can competently recognise actions across the two modalities including overlapping queries. 
Furthermore, we see consecutive actions are correctly recognised with varying interval lengths, such as the `open / close' audio actions between 0.2s and 4.5s. For AVE, TIM is able to distinguish between the background and a `barking' audio-visual event based on the time interval query. For the Perception Test, we see that TIM can distinguish between heavily overlapping actions across both modalities, such as between `break', `human speech', `hit' and `put something'. However, there are also failure cases, such as in EPIC-KITCHENS when the action `take washing up liquid' is recognised as `wash hand', as the model is likely confused by the context predominantly associated with the highly overlapping `wash hand' actions.

\subsection{Analysing Time Intervals}
\label{subsec:analyse_timeintervals}
We showcase the importance of effectively encoding time intervals, as well as how they differ from alternative strategies. We perform this analysis on the EPIC-KITCHENS-100 and EPIC-SOUNDS recognition tasks.

\begin{figure}[t]
  \centering
\includegraphics[width=\linewidth]{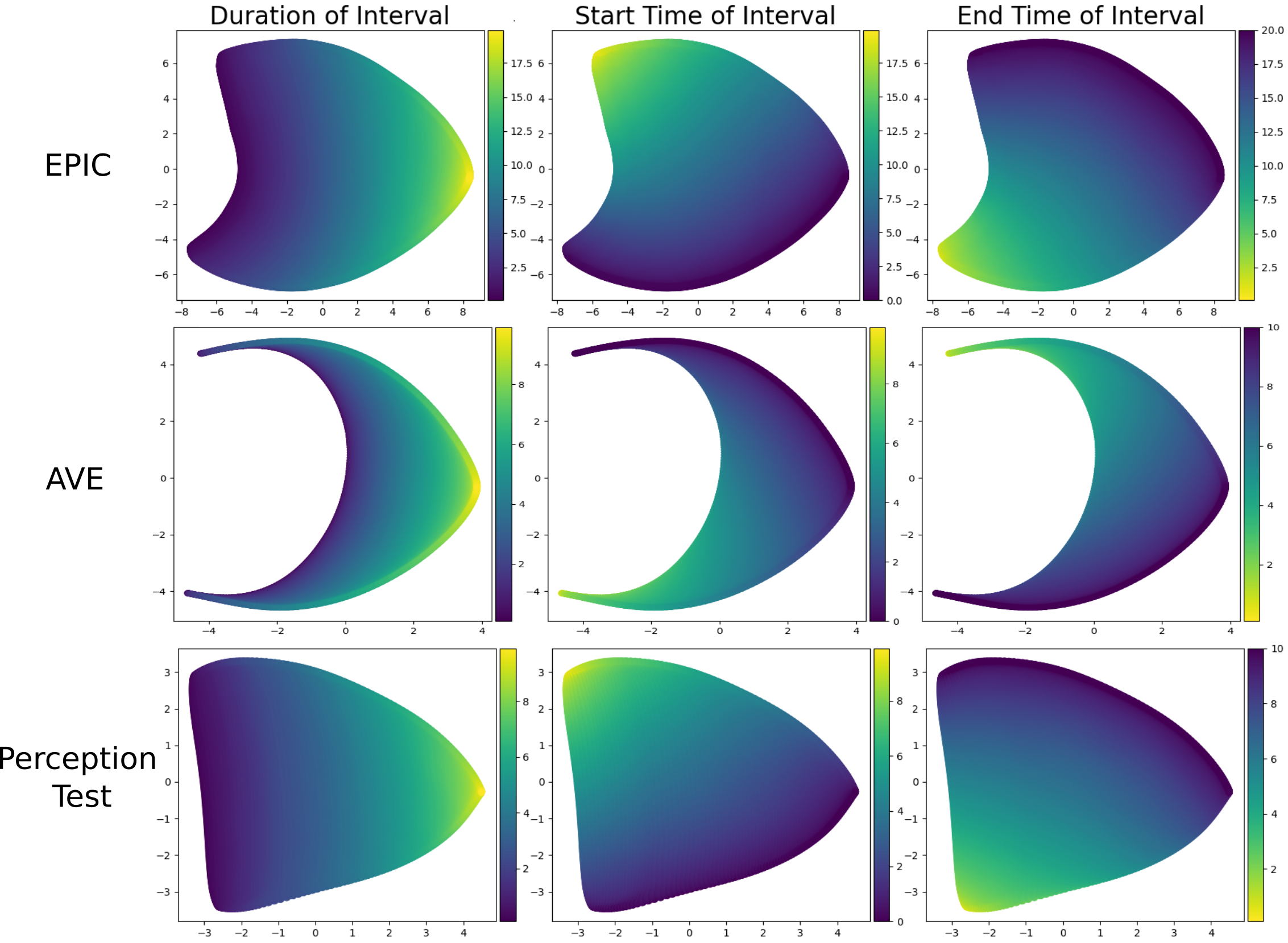}
  \caption{TSNE plot for time encodings $I(\cdot)$ on all datasets. In each plot, we use colour maps to indicate encodings of the time interval's duration (left), start time (middle) and end time (right).}
  \vspace*{-12pt}
  \label{fig:epic_tsne}
\end{figure}

\prepara\noindent\textbf{Time Encoding Representation.} To show TIM encodes time intervals across all datasets in Figure~\ref{fig:epic_tsne}. 
We use three colour maps on the same TSNE projection to show three properties of the encoded interval: duration, start time and end time. 
Interestingly, the 1D time encoding perfectly captures all three attributes and across the datasets.
While the encodings differ per dataset, as these differ in the positions and durations of actions, we see clear similarities in the learnt time encoding projections.
For example, the duration is perfectly captured across the x-axis of the TSNE plot with lower values indicating longer time intervals.

\begin{figure}[t]
  \centering
  \includegraphics[width=\linewidth]{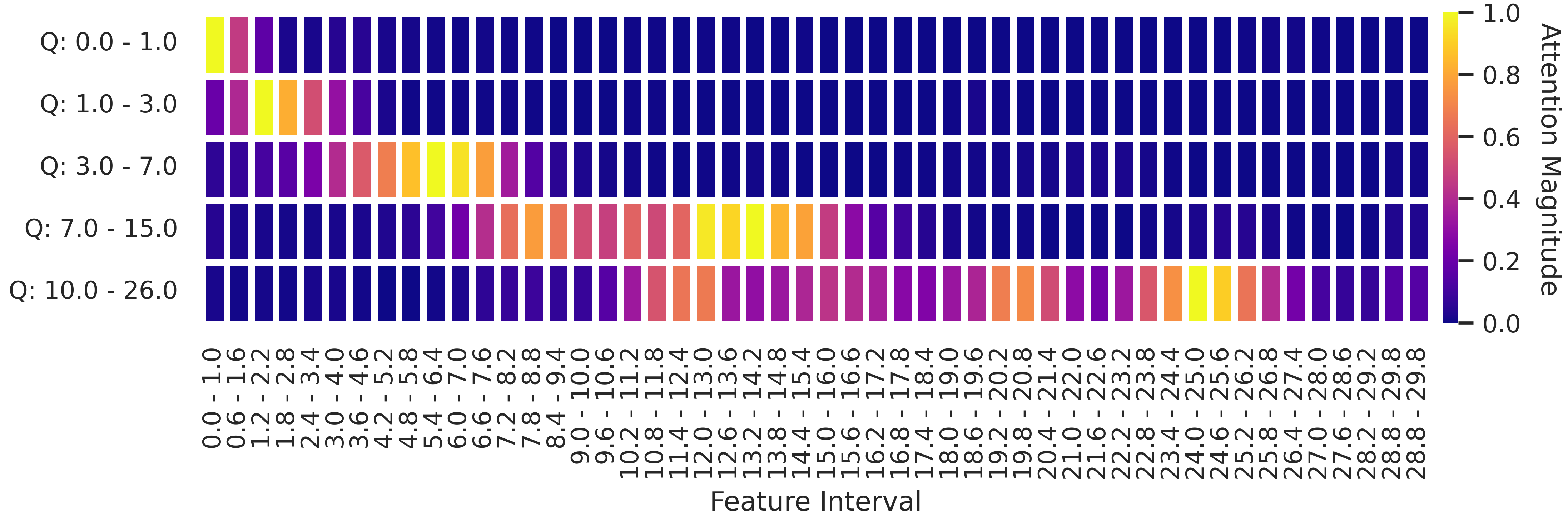}
  \includegraphics[width=\linewidth]{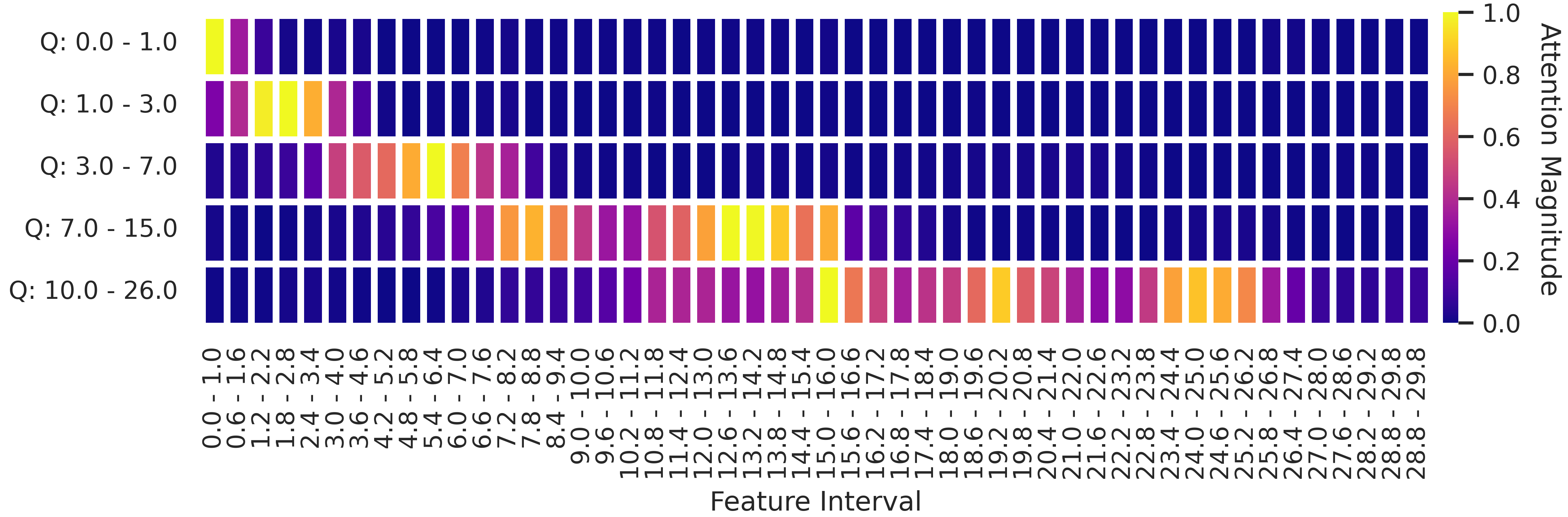}
  \vspace*{-12pt}
  \caption{Attention heatmaps of the second encoder layer for two random 30s clips in EPIC-KITCHENS. \textbf{x-axis}: input feature time intervals; \textbf{y-axis} query time intervals of varying position and duration. The attention magnitude relates to the query CLS token.}
  \label{fig:attn_heatmap}
\end{figure}

\prepara\noindent\textbf{Interval Query Attention.} We plot two attention heatmaps in Figure~\ref{fig:attn_heatmap} for 5 separate queries with varying positions and scales in EPIC-KITCHENS-100. 
We extract the attention weights from the second transformer encoder layer, as this appears to be the most relevant to the interval query. The learnt attention clearly applies to the feature time intervals contained within the query. 
We note the similarity between the attention in the two randomly selected windows.

\begin{figure}[t]
  \centering
  \includegraphics[width=0.8\linewidth]{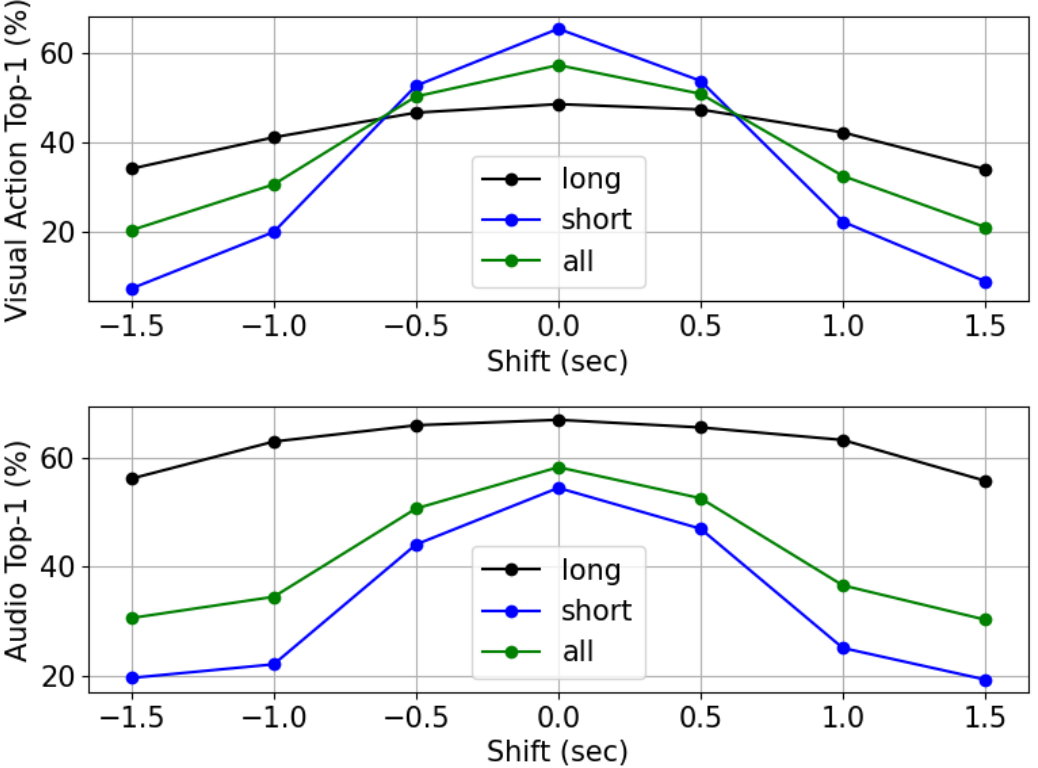}
  \caption{The impact of shifting the time interval query on visual performance (top) and audio performance (bottom), both on short actions ($<$ 2 sec) and long actions ($>$ 2 sec) and overall validation set  (\textbf{all}) for EPIC-KITCHENS-100 and EPIC-SOUNDS.}
  \vspace*{-15pt}
  \label{fig:sliding_window}
\end{figure}

\prepara\noindent\textbf{Shifting Intervals.}
To show how TIM effectively encodes the time interval of actions, we shift the time interval queries from their correct action interval by -1.5s to 1.5s, assessing the impact of these adjustments on the performance.

Figure~\ref{fig:sliding_window} shows the result.
We see the performance gradually drops, both in visual and audio, as the query interval moves away from the correct action interval.
The drop is also symmetric showcasing no bias.
Unsurprisingly, the performance drops significantly when shifting short actions both in video (-57.9\%) and audio (-35.2\%), while it is less extreme in long actions (-14.5\% and -11.2\%). We assess the impact of scaling the time interval in the ArXiv appendix.

\begin{table}[t]
\centering
\resizebox{1\linewidth}{!}{
\begin{tabular}{c c c c c }
\hline
& \multicolumn{3}{c}{\textbf{EPIC-KITCHENS}} & \textbf{EPIC-SOUNDS} \\
\hline
\textbf{Encoding} & \textbf{Verb} & \textbf{Noun} & \textbf{Action} & \textbf{Audio Actions} \\
\hline 
Learned & 43.8 & 44.3 & 29.6 & 23.7 \\
Sinusoidal & 43.8 & 44.6 & 30.0 & 13.4  \\
Centre & 74.3 & 65.8 & 55.6 & 56.4 \\
Separate-add & 76.0 & 66.2 & 56.4 & 57.7 \\
Interval-add & 76.3 & 66.5 & 56.9 & \textbf{58.8} \\
Separate-cat & 76.8 & \textbf{67.4} & 57.1 & 58.4 \\
Interval-cat (proposed) & \textbf{77.1} & 67.2 & \textbf{57.5} & 58.3 \\
\bottomrule
\end{tabular}}
\caption{Ablating the choice of encoding time intervals.}
\hfill
\vspace*{-15pt}
\label{tab:separateencoding}
\end{table}

\prepara\noindent\textbf{Time Interval Encodings.}
The Time Interval MLP encodes the interval of the query.
Here, we compare this to traditional positional encodings, both sinusoidal and learned. We also experiment on five different variations of the Time Interval MLP, namely:
(i) Centre -- we only encode the centre timestamp of the interval; (ii) Separate-Add/Cat -- we encode the intervals' start and end time separately and add the encoded output vectors together, or concatenate along the channel dimension; and (iii) Interval-Add/Cat -- we encode the start and end time within the same vector and add, or concatenate, the encoded output to the input sequence.

We show the results in Table~\ref{tab:separateencoding}. In all cases, the final encoding is of the same dimension for comparable results.
Performance is significantly worse with sinusoidal or learned positional encoding, as they are unable to capture the complexity of overlapping actions. 
There is also a drop when encoding only the centre of the time interval. 

Separate-Add/Cat are alternative ways to encode the intervals (and hence include duration information) resulting in comparable performance to the interval counterparts.
Our proposed approach to encoding the interval into the MLP shows the best performance for visual and while maintaining strong auditory performance.

\section{Conclusions}
\label{sec:conclusions}

In this paper, we propose to utilise the action's
time interval as a query to an audio-visual transformer
which learns to recognise the action from its interval and the unaltered surrounding context. We jointly train the model on modality-specific time intervals and label sets, allowing the Time Interval Machine (TIM) to recognise multiple events across both visual and auditory modalities. 

TIM is sensitive to the interval's position and duration. This allows the model, as is, to produce competitive results on action detection through multi-scale dense querying.

\noindent \textbf{Acknowledgements.} This work uses public datasets. It is supported by EPSRC Doctoral Training Program, EPSRC UMPIRE EP/T004991/1 and EPSRC Programme Grant VisualAI EP/T028572/1; and by the use of the EPSRC funded Tier 2 facility \mbox{JADE-II}.

{
    \small
    \bibliographystyle{ieeenat_fullname}
    \bibliography{main}
}

\appendix
\clearpage
\setcounter{page}{1}
\maketitlesupplementary
\appendix
\part{Appendix} %
\parttoc %

\section{Further analysis of time intervals -- scaling}
\label{sec:scale_intervals}
\begin{figure}[t]
  \centering
\includegraphics[width=\linewidth]{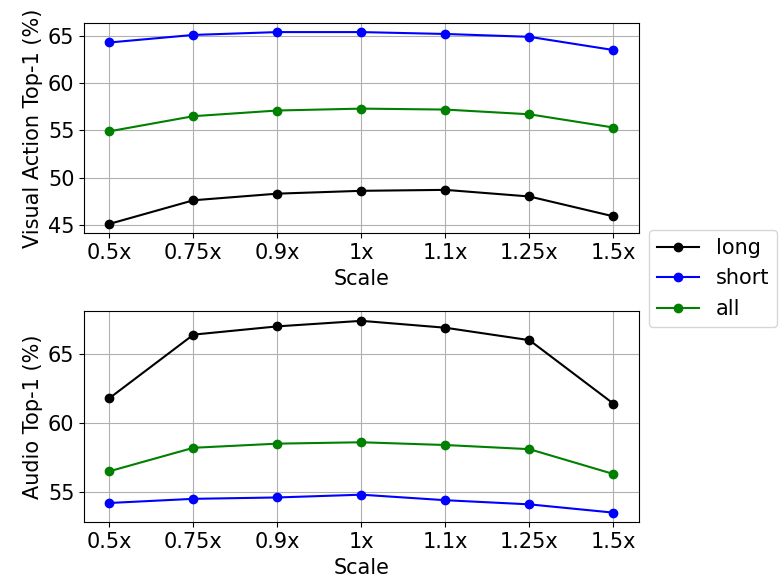}
  \caption{The impact of scaling a query centered around the action on both visual and audio performance. A shift of 0.0 sec, or a 1x scale, means querying with original time interval. Both visual and audio performance fall gradually as the query moves away from the original time interval. }
  \label{fig:scaling_window}
\end{figure}

We show the effect of shifting the time interval queries from their correct action interval in Section~\ref{subsec:analyse_timeintervals} (Figure~\ref{fig:sliding_window}). 
In Figure~\ref{fig:scaling_window}, we show the analogous figure as we vary the effect of \textit{scaling} a centralised query from the ground truth. Similar to shifting, we also demonstrate a decrease in performance when scaling a query. Performance drops from 57.5\% to 54.9\% when contracting, and to 55.3\% when expanding the query in visual queries. In audio, we see a drop from 58.3\% to 56.5\% when contracting, and to 56.3\% when expanding the query.

Both Figure~\ref{fig:sliding_window} and Figure~\ref{fig:scaling_window} combined showcase the ability of TIM to correctly model the time interval of actions. The performance drops steadily, yet smoothly, as queries are changed from the ground truth -- whether shifted or scaled.

\section{Test Set Results}
In this section we showcase TIM's results on multiple challenges and test sets across EPIC, namely EPIC-KITCHENS-100 recognition, EPIC-Sounds recognition, EPIC-KITCHENS-100 detection and EPIC-Sounds detection. 

\subsection{EPIC-KITCHENS-100 Test Set}
\label{sec:epic_test_set}
In the main paper, TIM is evaluated on the EPIC-KITCHENS-100 validation set, as most state-of-the-art results only report on the validation set, and thus we do the same for a direct comparison. Here, we evaluate the same model on the test set, by submitting to the leaderboards.

\begin{figure}
    \centering
    \includegraphics[width=1\linewidth]{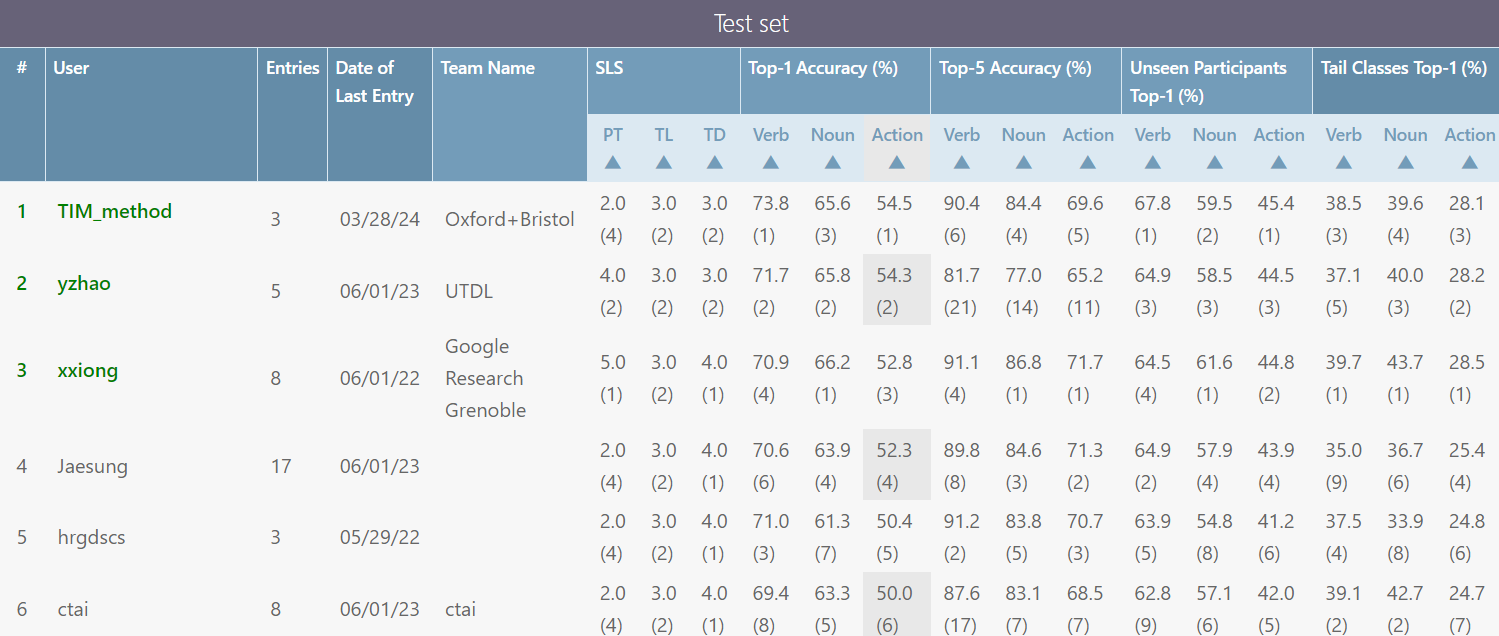}
    \caption{Screenshot of the EPIC-KITCHENS Action Recognition leaderboard (March 2024) showcasing our TIM\_method ranked top.}
    \label{fig:recognition-leaderboard}
\end{figure}

We report the results of our best performing model in Table~\ref{tab:sota_epic_test}.
We ensemble six TIM models with input window lengths $W = 15, 30, 36, 40, 45, 60$ seconds with weights $[1.0, 0.9, 0.9, 0.9, 0.9, 0.9]$ respectively. All other parameters/architecture details remain unchanged.
Our model achieves SOTA action performance (which ranks the winners) as well as verb performance. 
TIM remains behind SOTA on noun performance by 0.6\%.
We also report a single model TIM, without ensembling, and show this is competitive with winners from previous years despite using only a single model. We showcase the submission ranked top on the test set leaderboard in Figure~\ref{fig:recognition-leaderboard}.

\begin{figure}
    \centering
    \includegraphics[width=1\linewidth]{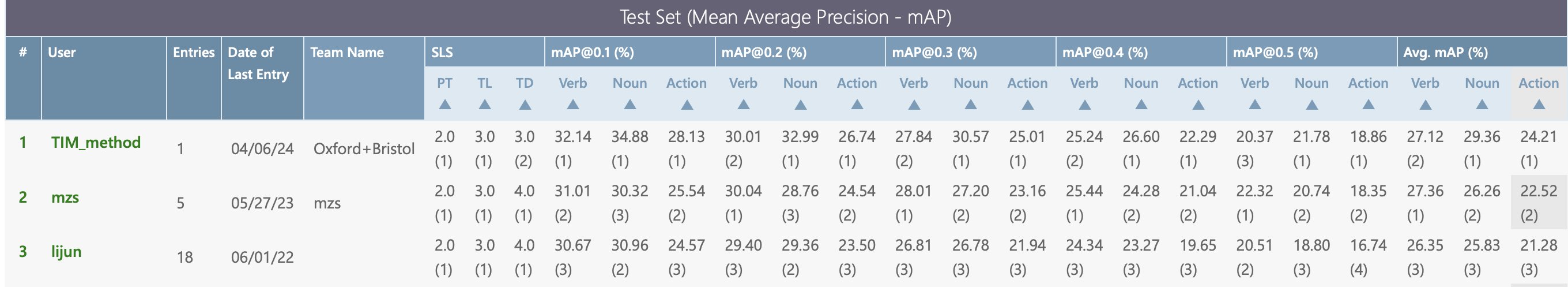}
    \caption{Screenshot of the EPIC-KITCHENS Action detecion leaderboard (April 2024) showcasing our TIM\_method ranked top.}
    \label{fig:detection-leaderboard}
\end{figure}

We also provide results for detection in Table~\ref{tab:sota_epic100_test_det}. Note that this challenge also requires action predictions i.e.\ a combination of a verb prediction and noun prediction. To achieve this, we combine the predictions of each query from our verb and noun model, resulting in a two-stream architecture. We then follow~\cite{zhang2022actionformer} and re-weight the confidence and action boundaries of each proposal by:

\vspace{-6pt}
\begin{equation}
\begin{split}
    \mathbf{p}(action)\ &=\ \mathbf{p}(verb)^{\alpha} \ \mathbf{p}(noun)^{(1-\alpha)}, \\
    \mathbf{d}(action)\ &=\ \omega \mathbf{d}(verb) +  (1-\omega) \mathbf{d}(noun) \\
\end{split}
\end{equation}
\vspace{-6pt}

where $\alpha\! = \! 0.45$ and $\omega = \mathbf{p}(verb) / (\mathbf{p}(verb)+\mathbf{p}(noun))$. We can see that TIM sets a new SOTA in noun and action detection by 3.1 and 1.7 mAP respectively, only falling slightly behind on verb. For this method we ensemble 6 models using context windows $W = 15, 30, 45$ for both verb and noun streams. Evidence of our new SOTA method is shown in Figure~\ref{fig:detection-leaderboard}.

\begin{table}[t]
\centering
\begin{tabular}{l c c c c }
\hline
\textbf{Method} & \textbf{Ensemble} & \textbf{Verb} & \textbf{Noun} & \textbf{Action} \\
\hline
ctai & \checkmark & 69.4 & 63.3 & 50.0 \\
hrgdscs & \checkmark & 71.0 & 61.3 & 50.4 \\
Jaesung & \checkmark & 70.6 & 63.9 & 52.3 \\
xxiong & \checkmark & 70.9 & \textbf{66.2} & 52.8 \\
\rowcolor{LightGrey} \textbf{TIM (ours)} & \xmark & 73.1 & 64.1 & 53.0 \\
yzhao & \checkmark & 71.7 & 65.8 & 54.3 \\
\rowcolor{LightGrey} \textbf{TIM (ours)} & \checkmark & \textbf{73.8} & 65.6 & \textbf{54.5} \\
\hline 
\end{tabular}
\caption{Comparisons to state-of-the-art {\em recognition} models on EPIC-KITCHENS test set. We report the top-1 accuracy for verb, noun and action (\%).}
\label{tab:sota_epic_test}
\end{table}

\begin{table}[t]
    \centering
    \resizebox{\linewidth}{!}{
        \begin{tabular}{| c | c | c c c c c | c|}            \hline
             \multirow{2}{*}{\textbf{Method}} & \multicolumn{7}{c|}{\textbf{Average Precision (AP)}} \\
             \cline{2-8} & Task & @0.1 & @0.2 & @0.3 & @0.4 & @0.5 & Avg. \\
             \hline 
            \multirow{3}{*}{lijun} & Verb & 30.7 & 29.4 & 26.8 & 24.3 & 20.5 & 26.4 \\
            & Noun & 31.0 & 29.4 & 26.8 & 23.3 & 18.8 & 25.8 \\
            & Action & 24.6 & 23.5 & 21.9 & 19.7 & 16.7 & 21.3 \\
            \hline
            \multirow{3}{*}{mzs} & Verb & 31.1 & 28.0 & 26.5 & \textbf{25.4} & \textbf{22.3} & \textbf{27.3} \\
            & Noun & 30.3 & 28.8 & 27.2 & 24.3 & 20.7 & 26.3 \\
            & Action & 25.5 & 24.5 & 23.2 & 21.0 & 18.4 & 22.5 \\
            \hline
            \multirow{3}{*}{TIM} & Verb & \textbf{32.1} & \textbf{30.0} & \textbf{27.8} & 25.2 & 20.4 & 27.1 \\
            & Noun & \textbf{34.9} & \textbf{33.0} & \textbf{30.6} & \textbf{26.6} & \textbf{21.8} & \textbf{29.4} \\
            & Action & \textbf{28.1} & \textbf{26.7} & \textbf{25.0} & \textbf{22.3} & \textbf{18.9} & \textbf{24.2} \\
            \hline
        \end{tabular}
    }
    \caption{Comparisons to state-of-the-art {\em visual action detection} models on the EPIC-KITCHENS test set. We report the average precision at IOU thresholds $[0.1, 0.2, 0.3, 0.4, 0.5]$ as well as their average across all thresholds on verb, noun and action.}
    \label{tab:sota_epic100_test_det}
\end{table}

\subsection{EPIC-Sounds Test Set}
\begin{figure}
    \centering
    \includegraphics[width=1\linewidth]{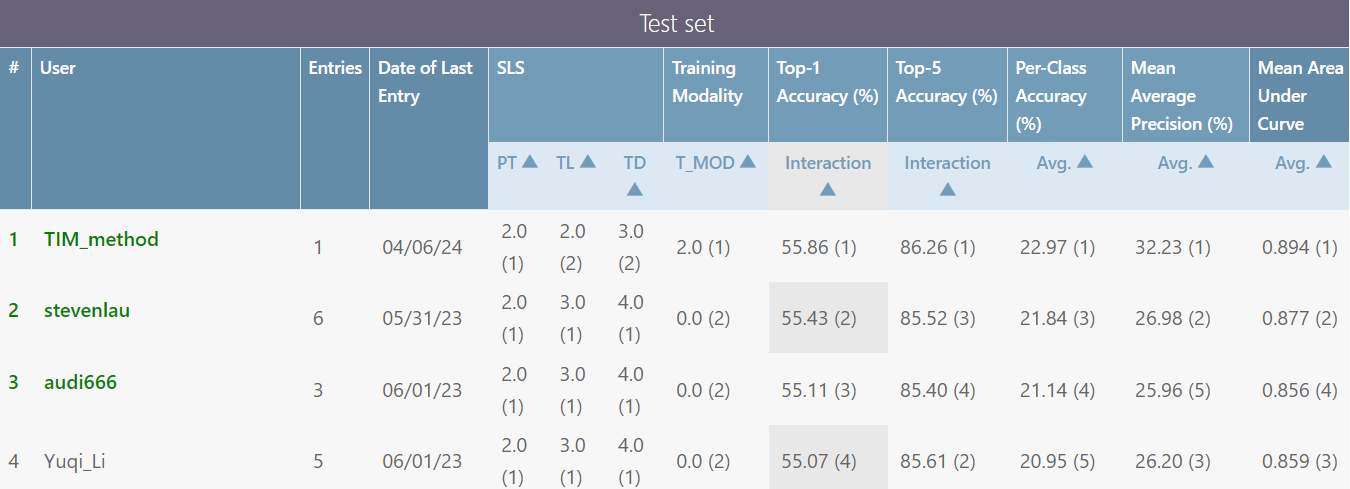}
    \caption{Screenshot of the EPIC-Sounds Audio-Based Interaction Recognition leaderboard (April 2024) showcasing our TIM\_method ranked top.}
    \label{fig:sounds-recognition-leaderboard}
\end{figure}

Here, we evaluate TIM on the test set by submitting to leaderboards. Again, we showcase results for both single and ensemble models in Table~\ref{tab:sota_epic_sounds_test}, using the same configuration described previously in the EPIC-KITCHENS-100 Action Recognition Challenge. Our model achieves SOTA performance across all metrics. Our single model does not perform quite as well as the visual counterpart in top-1 accuracy, but still outperforms all other methods with regards to mean average precision and per-class accuracy. It is also worth noting that the model selection was visually biased i.e.\ we chose the best performing visual model, instead of audio. Again, we showcase the submission ranked top on the test set leaderboard in Figure~\ref{fig:sounds-recognition-leaderboard}.

We also provide detection results in Table~\ref{tab:sota_epic_sounds_test_det}, where we convincingly outperform the ActionFormer baseline across all metrics, notably by 4.2 mAP, setting a new SOTA in this challenge.

\begin{table}[t]
\centering

\resizebox{\linewidth}{!}{
\begin{tabular}{l c c c c}
\hline
\textbf{Method} & \textbf{Ensemble} & \textbf{Top-1 Acc.} & \textbf{PCA} & \textbf{mAP}
\\
\hline
\rowcolor{LightGrey} \textbf{TIM (ours)} & \xmark & 54.9 & 22.8 & 31.9 \\
Yuqi\_Li & \checkmark & 55.1 & 21.0 & 26.2 \\
audi666 & \xmark & 55.1 & 21.1 & 26.0 \\
stevenlau & \xmark & 55.4 & 21.8 & 27.0 \\
\rowcolor{LightGrey} \textbf{TIM (ours)} & \checkmark & \textbf{55.9} & \textbf{23.0} & \textbf{32.2} \\
\hline 
\end{tabular}
}
\caption{Comparisons to state-of-the-art {\em audio recognition} models on EPIC-Sounds test set. We report the top-1 accuracy for audio interactions, along with the per-class accuracy (PCA) and mean average precision (mAP).}
\label{tab:sota_epic_sounds_test}
\end{table}

\begin{table}[t]
    \centering
    \resizebox{\linewidth}{!}{
        \begin{tabular}{| c | c c c c c | c|}            \hline
             \multirow{2}{*}{\textbf{Method}} & \multicolumn{6}{c|}{\textbf{Average Precision (AP)}} \\
             \cline{2-7} & @0.1 & @0.2 & @0.3 & @0.4 & @0.5 & Avg. \\
             \hline 
            ActionFormer Baseline & 9.6 & 8.5 & 7.4 & 6.2 & 5.1 & 7.4 \\
            \hline
            TIM & \textbf{15.7} & \textbf{13.3} & \textbf{11.4} & \textbf{9.3} & \textbf{7.3} & \textbf{11.4} \\
            \hline
        \end{tabular}
    }
    \caption{Comparisons to state-of-the-art {\em audio detection} models on the EPIC-Sounds test set. We report the average precision at IOU thresholds $[0.1, 0.2, 0.3, 0.4, 0.5]$ as well as their average across all thresholds.}
    \label{tab:sota_epic_sounds_test_det}
\end{table}

\begin{figure}
    \centering
    \includegraphics[width=1\linewidth]{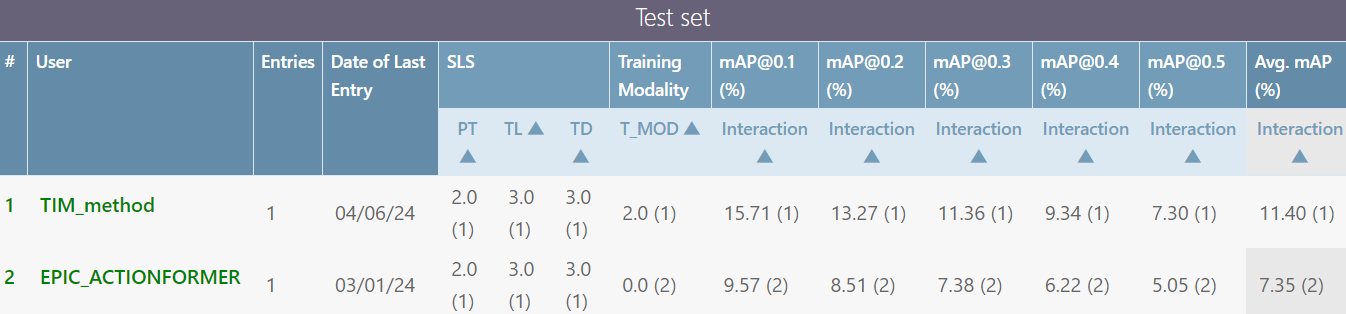}
    \caption{Screenshot of the EPIC-Sounds Audio-Based Interaction Detection leaderboard (April 2024) showcasing our TIM\_method ranked top.}
    \label{fig:sounds-detection-leaderboard}
\end{figure}

\section{Ablation studies}
\label{sec:ablation}
This section contains ablation studies on the proposed TIM architecture in various aspects and loss functions. We perform all ablations on the EPIC-KITCHENS (visual action recognition) and EPIC-SOUNDS (audio actions recognition). In all tables, we highlight our main reported results in grey.

\prepara\noindent\textbf{Number of encoder layers.}
Here, we ablate the number of transformer encoder layers  
in TIM, varying from 1 to 6, on the performance. 
Other hyperparameters and model configuration remain fixed, as described in our main paper.
Table~\ref{tab:ab_encoderlayers} shows the result. 

The best visual action performance is obtained by using four layers, while verb and noun performance is comparable to the models with only three layers. 
Interestingly, audio performance is best when using three layers.
This is likely due to overfitting of the audio input compared to the visual. It is well known that multimodal training is susceptible to differences between the two modalities~\cite{Wang_2020_CVPR}.
However, our training regime remains relatively stable between the two modalities. 
The difference between the top performance audio (3 layer) and our reported results (4 layers) is only~1.0\%.

\begin{table}[t]
\centering
\resizebox{1\linewidth}{!}{
\begin{tabular}{c c c c c }
\hline
& \multicolumn{3}{c}{\textbf{EPIC-KITCHENS}} & \textbf{EPIC-SOUNDS} \\
\hline
\textbf{Depth} & \textbf{Verb} & \textbf{Noun} & \textbf{Action} & \textbf{Audio Actions} \\
\hline 
1 Layer  & 75.8 & 65.0 & 55.4 & 58.4 \\
2 Layers & 76.5 & 66.2 & 56.5 & 58.4 \\
3 Layers & 77.0 & 66.9 & 57.2 & \textbf{59.3} \\
\rowcolor{LightGrey} 4 Layers & \textbf{77.1} & \textbf{67.2} & \textbf{57.5} & 58.3 \\
5 Layers& 76.6 & 66.7 & 56.9 & 58.2 \\
6 Layers& 76.9 & 66.6 & 57.0 & 57.5 \\
\end{tabular}}
\caption{Effect of changing the number of transformer encoder layers. 
Number of transformer head is fixed to 16. The highlighted row is the performance we report in the main paper.}
\hfill
\label{tab:ab_encoderlayers}
\end{table}

\prepara\noindent\textbf{Number of transformer heads.}
We also ablate the number of transformer heads.
We experiment with 2, 4, 8 and 16, keeping other hyperparameters fixed.
Table~\ref{tab:ab_head} shows the results of this ablation.

Peak visual and audio performance is obtained when using 8 heads. This is the performance we report in the main paper. Interestingly, changing the number of heads has a comparable impact on performance to that when changing the number of layers reported in Table~\ref{tab:ab_encoderlayers}.

\begin{table}[t]
\centering
\resizebox{1\linewidth}{!}{
\begin{tabular}{c c c c c }
\hline
& \multicolumn{3}{c}{\textbf{EPIC-KITCHENS}} & \textbf{EPIC-SOUNDS} \\
\hline
\textbf{\# Head} & \textbf{Verb} & \textbf{Noun} & \textbf{Action} & \textbf{Audio Actions} \\
\hline 
2 & 77.0 & 65.9 & 56.6 & \textbf{58.3} \\
4 & 76.7 & 66.7 & 56.9 & 57.9 \\
\rowcolor{LightGrey} 8 & \textbf{77.1} & \textbf{67.2} & \textbf{57.5} & \textbf{58.3} \\
16 & 77.0 & \textbf{67.2} & 57.1 & 58.1 \\
\end{tabular}}
\caption{Effect of changing the number of heads in transformer. Number of transformer layer is fixed to 4. The highlighted row is the performance we report in the main paper.}
\hfill
\label{tab:ab_head}
\end{table}

\prepara\noindent\textbf{Temporal distance regression head architecture.}
We also ablate the structure of the Temporal distance regression head $h_{\tilde{t}}$ in Eq~\ref{eq:aux_loss}, by varying the number of layers from 1 to 4.
Results are shown in Table~\ref{tab:ab_auxhead}.
The results are similar across all depths, but we find using 3 layers gives the best compromise across all metrics and these are the results we report in the paper.

\begin{table}[t]
\centering
\resizebox{1\linewidth}{!}{
\begin{tabular}{c c c c c }
\hline
& \multicolumn{3}{c}{\textbf{EPIC-KITCHENS}} & \textbf{EPIC-SOUNDS} \\
\hline
\textbf{Depth} & \textbf{Verb} & \textbf{Noun} & \textbf{Action} & \textbf{Audio Actions} \\
\hline 
1 Layer & 77.0 & 66.8 & 57.3 & 58.1 \\
2 Layers & \textbf{77.2} & 66.9 & 56.9 & 58.4 \\
\rowcolor{LightGrey} 3 Layers & 77.1 & \textbf{67.2} & \textbf{57.5} & 58.3 \\
4 Layers & 76.8 & 66.9 & \textbf{57.5} & \textbf{58.7} \\
\end{tabular}}
\caption{Effect of temporal distance head structure. The highlighted row is the performance we report in the main paper.}
\hfill
\label{tab:ab_auxhead}
\end{table}

\prepara\noindent\textbf{Input length and feature density.}
We set $W\geq10$ seconds. These long segments from untrimmed videos are complex and contain multiple overlapping actions. For example in EPIC-KITCHENS-100, a 30 second window contains on average 16 audio-visual annotated events with a maximum of 81 queries in the training set. Additionally, 28.1\% of all actions overlap. 

Table~\ref{tab:num_feats} shows the effect of changing the input visual and audio features for TIM. We experiment with the window size $W$, which is affected by the number of features within the window ($N^{m}$) and the stride between the features ($H_{f}$). We also experiment with the window stride ($H_{w}$) which affects how many windows fit within an entire untrimmed video, and hence the extent of the temporal context around a given action.
We separate the table into 4 sections, separated by horizontal lines, to showcase different variations.

First we ablate on the number of features, while keeping the feature hop size constant. 
Increasing the number of features will increase the window size. 
We see that using 50 features with a stride of 0.6 seconds works best, resulting in a window size of 30 seconds. 
This time frame likely provides enough relevant context to the action without injecting redundant information through additional features too distant from the action.

We then ablate the feature stride while keeping the number of features constant. 
In this case, a larger hop size results in a larger input window. 
We see that a stride of 0.6 seconds, resulting in a 30 second window, performs the best. 
This outperforms a 30 second window with 75 features with stride 0.4 seconds, as the sparser sampling likely removes redundant information.

We also experiment with the feature density, by fixing the window size to 30 seconds, but varying both the number of features and the feature stride. 
In this case, we see that our proposed feature density of $N^{m}=50$ performs best. 
Increasing the number of features increases redundancy, whereas a sparser number does not benefit from sufficient neighbouring context.

Finally, we experiment on the stride of the input window. 
A smaller stride results in increased overlap between the input features.
Compared to the stride of 1.0, used in our results, an increased stride clearly drops visual performance. 

\begin{table}[t]
    \centering
    \resizebox{\linewidth}{!}{
        \begin{tabular}{c c c c c c c c }
            \hline
            & & & & \multicolumn{3}{c}{\textbf{EPIC-KITCHENS}} & \textbf{EPIC-SOUNDS} \\
            \hline
            \textbf{W} & $\mathbf{N^{m}}$ & $\mathbf{H_{f}}$ & $\mathbf{H_{w}}$ & \textbf{Verb} & \textbf{Noun} & \textbf{Action} & \textbf{Audio Actions} \\
            \hline
            15.0 & 25 & 0.6 & 1.0 & 76.8 & 67.0 & 57.3 & \textbf{59.0} \\
            45.0 & 75 & 0.6 & 1.0 & 76.6 & 67.1 & 57.0 & 57.4 \\
            60.0 & 100 & 0.6 & 1.0 & 76.5 & 66.8 & 57.1  & 57.3 \\
            \hline
            10.0 & 50 & 0.2 & 1.0 & 76.2 & 66.1 & 55.9 & 58.4 \\
            20.0 & 50 & 0.4 & 1.0 & 76.7 & 66.7 & 56.8 & 58.7 \\
            \rowcolor{LightGrey} 30.0 & 50 & 0.6 & 1.0 & \textbf{77.1} & \textbf{67.2} & \textbf{57.5} & 58.3 \\
            40.0 & 50 & 0.8 & 1.0 &  76.5 & 66.8 & 56.8 & 58.0 \\
            50.0 & 50 & 1.0 & 1.0 &  75.5 & 65.9 & 56.2 & 56.5 \\
            \hline 
            30.0 & 25 & 1.2 & 1.0 & 76.5 & 66.1 & 56.4 & 57.3 \\
            30.0 & 75 & 0.4 & 1.0 & 76.8 & 66.5 & 57.3 & 58.0 \\
            \hline
            30.0 & 50 & 0.6 & 2.0 & 76.7 & 66.8 & 57.2 & 58.7 \\
            30.0 & 50 & 0.6 & 5.0 & 76.4 & 66.1 & 56.4 & 58.6 \\
            30.0 & 50 & 0.6 & 10.0 & 75.5 & 65.4 & 55.6 & 57.6 \\
            \hline 
        \end{tabular}
    }
    \caption{Effect of changing the parameters to alter the feature input to TIM in EPIC-KITCHENS and EPIC-SOUNDS.  \textbf{W}: Window Size in seconds, $\mathbf{N^{m}}$: Number of features,  $\mathbf{H_{f}}$: Feature Stride in seconds, $\mathbf{H_{w}}$: Window Stride in seconds.}
    \label{tab:num_feats}
\end{table}

\prepara\noindent\textbf{Time Interval MLP structure.} 
We also ablate the structure of the Time Interval MLP $I(.)$. 
We experiment with varying the number of linear layers.
As shown in Table~\ref{tab:time_mlp}, TIM seems to favour a depth of 3 within the Time Interval MLP, benefiting from a 1.0\% visual and 0.5\% audio boost over 2 layers.

\begin{table}[t]
\centering
\resizebox{0.9\linewidth}{!}{
\begin{tabular}{c c c c c }
\hline
& \multicolumn{3}{c}{\textbf{EPIC-KITCHENS}} & \textbf{EPIC-SOUNDS} \\
\hline
\textbf{Depth} & \textbf{Verb} & \textbf{Noun} & \textbf{Action} & \textbf{Audio Actions} \\
\hline 
1 Layer & 75.5 & 66.3 & 56.0 & 57.2 \\
2 Layers & 76.6 & 66.5 & 56.5 & 57.8 \\
\rowcolor{LightGrey} 3 Layers & \textbf{77.1} & \textbf{67.2} & \textbf{57.5} & 58.3 \\
4 Layers & 76.5 & 66.9 & 57.3 & 58.0 \\
5 Layers & 76.6 & 67.0 & 57.2 & \textbf{58.4} \\
\end{tabular}
}

\caption{Effect of the Time Interval MLP $I(\cdot)$ structure. The highlighted row is the performance we reported in the main paper.}
\hfill
\label{tab:time_mlp}
\end{table}

\prepara\noindent\textbf{Loss ablation.} 
We experiment with varying the $\lambda^{td}$, as well as the within-modal and cross-modal sampling variants (Eq.~\ref{eq:aux_loss}). 
The results are shown in Table~\ref{tab:aux_loss}.
Introducing the Temporal Distance loss ($\lambda^{td} > 0$) improves the overall performance for visual, but has an adverse effect on audio.
We also observe that $\lambda^{td}=0.3$ with cross-modal sampling shows the highest performance on the visual action metric. The
\textit{cross-modal} sampling strategy shows marginally improved results than the \textit{within-modal} strategy for visual, suggesting the distance loss is more beneficial for video than for audio.

\begin{table}[t]
\centering
\resizebox{\linewidth}{!}{
\begin{tabular}{c c c c c c }
\hline
& & \multicolumn{3}{c}{\textbf{EPIC-KITCHENS}} & \textbf{EPIC-SOUNDS} \\
\hline
$\lambda^{td}$ & \textbf{Sampling} & \textbf{Verb} & \textbf{Noun} & \textbf{Action} & \textbf{Audio Actions} \\
\hline 
0.0 & - & 76.9 & 66.7 & 57.2 & \textbf{58.4} \\
0.1 & \textit{cross-modal} & 77.0 & 66.7 & 57.1 & 58.1 \\
\rowcolor{LightGrey} 0.3 & \textit{cross-modal} & 77.1 & \textbf{67.2} & \textbf{57.5} & 58.3 \\
0.3 & \textit{within-modal} & \textbf{77.3} & 67.0 & 57.4 & \textbf{58.4} \\
0.5 & \textit{cross-modal} & 76.9 & 66.8 & 57.3 & 58.2 \\

\end{tabular}}
\hfill
\caption{Effect of Temporal Distance loss on performance. \textbf{Sampling} represents the two different ways of sampling pairs $\mathbb{B}$, \textit{cross-modal} means sampling the pairs across modalities and \textit{within-modal} indicates sampling pairs only within the same modality. We report the highlighted row in the main paper.}
\label{tab:aux_loss}
\end{table}

\section{TIM for Detection}
\label{sec:tim_for_det}
In this section we describe how we adapted TIM for the task of action detection for the results reported in Table~\ref{tab:sota_epic100_det}. The backbone remains largely unchanged from the recognition task. However, there are differences in how we obtain queries, as well as an additional interval regression head.

\subsection{Multi-scale Queries for Detection}
While in recognition we can utilise the ground-truth timestamps of the actions to query the input, in detection, we obtain dense {\em proposal} queries by constructing a query pyramid. 
These queries cover multiple fixed-size scales, spanning the entirety of the long video at each level, starting from short, dense temporal interval queries to long ones. 
The pyramid structure allows the model to classify and regress to both long and short actions within the input. 

In practice, when constructing our query pyramid, we start from a query interval size of $0.005 * W$ (0.15s for a $W=30$s window), with dense queries that span the entire window. We then double the query size in the next layer, again spanning the entire window at this resolution, and repeat this process stopping before the query size matches or exceeds the full window size. For a 30s window, this method constructs a query pyramid consisting of 8 layers, with resolutions: [0.15s, 0.3s, 0.6s, 1.2s, 2.4s, 4.8s, 9.6s, 19.2s].

We classify these queries in the same manner as recognition. However, we also introduce a regression head, which predicts the start and end times of the action the query is assigned to. The regression head allows for temporal localisation to be improved over that of the proposal interval and to have greater overlap with the ground truth.

When obtaining the final sets of detections, we classify and regress all queries in the pyramid across all input windows in the untrimmed video. We then threshold predictions that are below a confidence threshold. 
We then apply class-dependent Soft-NMS~\cite{bodla2017softnms} to the filtered predictions to remove highly overlapping proposals, before calculating the precision scores.

\subsection{Detection Training}
During training, we deem any query in the fixed pyramid (multi-scale) set of queries with temporal $IOU \geq 0.6$ with any ground truth action as a positive query.
If a query has a temporal overlap above the threshold with multiple ground-truth actions, we only consider the action label with the highest $IOU$. 
For all positive queries, we directly predict the assigned action's start and end times $(t^{m}_{s}, t^{m}_{e})$ and classify the corresponding action label.
For negative queries, we do not regress the interval's duration and set the label as a zero-vector for across all classes e.g.\ background. 

As with recognition, we classify all queries with $h^{m}_{\texttt{CLS}}(\cdot)$ and obtain predictions $\hat{y}^{m}_{\texttt{CLS}} = h^{m}_{\texttt{CLS}}(Z^{m}_{\texttt{CLS}})$. 
To classify queries, we train TIM using a Sigmoid Focal Loss~\cite{lin2020focalloss} $F(\cdot)$ to balance the positive and negative samples:
\begin{equation}
  L^{m}_{\texttt{det\_CLS}} = \frac{1}{B}\sum^{B} F (\hat{y}^{m}_{\texttt{CLS}}, y^{m}_{\texttt{CLS}})
  \label{eq:cls_det_loss}
\end{equation}
For positive queries, we also feed the encoded $\texttt{CLS}$ tokens through a separate regression head $h^{m}_{\texttt{REG}}$ to predict the querys associated ground truth action start and end time $(\hat{t}^{m}_{s}, \hat{t}^{m}_{e}) = h^{m}_{\texttt{REG}}(Z^{m}_{\texttt{CLS}})$. 
We train this via a DIOU regression loss~\cite{zheng2020diou}:
 \begin{equation}
  L^{m}_{\texttt{det\_REG}} =\frac{1}{Q_{P}}\sum^{Q_{P}} DIOU((\hat{t}^{m}_{s}, \hat{t}^{m}_{e}), (t^{m}_{s}, t^{m}_{e}))
  \label{eq:reg_det_loss}
\end{equation}
where $Q_{P}$ is the number of positive queries. 
Finally, we combine both losses to form our detection loss:
 \begin{equation}
  L^{m}_{det} = L^{m}_{\texttt{det\_CLS}} + \lambda_{\texttt{det\_REG}} L^{m}_{\texttt{det\_REG}}
  \label{eq:detection_loss}
\end{equation}

Where $\lambda_{\texttt{det\_REG}}$ is a parameter used to weight the regression loss. We set this to $0.5$.

\section{Further Implementation Details}
\label{sec:further_imp}

\prepara\noindent\textbf{Feature Extraction} The Omnivore model used is pre-trained with ImageNet~\cite{russakovsky2015imagenet}, Kinetics~\cite{kay2017kinetics} and SUN RGB-D~\cite{song2015sun} datasets.
For EPIC experiments, we finetune the model with EPIC-KITCHENS100 visual labels.
The VideoMAE-L features are pre-trained on Kinetics~\cite{kay2017kinetics}, Something-Something V2~\cite{goyal2017something}, AVA~\cite{gu2018ava} and WebVid2M, which we also fine-tuned on EPIC-KITCHENS visual labels. 
The detailed training procedure for Omnivore is available in~\cite{girdhar2022omnivore} and for VideoMAE in~\cite{tong2022videomae, wang2022internvideo}.
We extract dense features that overlap, so that we can use fine-grained time intervals as a query.
Each 1 second Omnivore feature is computed by feeding 32 frames the temporal segment sampling described in~\cite{liu2022video} and whereas we feed 16 frames using the sampling described in~\cite{tong2022videomae,wang2022internvideo} for each VideoMAE feature. 

For Auditory SlowFast~\cite{kazakos2021slow}, we utilise the pre-trained VGGSound~\cite{chen2020vggsound} model and change the input length from 2 seconds to 1 second to match the temporal extent of the visual features.
Only for EPIC experiments, we finetune the model with EPIC-SOUNDS audio labels.
The additional sets used for data augmentation apply SpecAugment with two frequency masks with F = 27 and two time masks with T = 25. Again, this enables data augmentation for audio.

For AVE visual features, we use a VGG-19~\cite{simonyan2014very} model pre-trained on ImageNet~\cite{russakovsky2015imagenet}.
We extract the features from \textit{pool5} layer on VGG-19 to get a spatial feature map per each frame.
We average these feature maps per each second by global pooling.
For audio features, we adopt a VGG-like~\cite{hershey2017cnn} network pre-trained on AudioSet~\cite{gemmeke2017audio}.
Both visual and audio feature cover one second of the visual or audio stream. Additionally, due to the significantly smaller size of the AVE dataset, we reduce the model size for this dataset by halving the hidden dimension of all linear layers ($512$-D) and applying channel-wise dropout with $p=0.1$ to the raw input features, but retain a dropout of $p=0.5$ on the encoded transformer input. 

\prepara\noindent\textbf{Model selection scheme.} For datasets with distinct visual and audio label sets (EPIC and Perception Test), we train a single model on both sets of labels simultaneously. In these cases, we report results across all metrics in both modalities for the epoch with the best visual performance.
We note that we can obtain additional audio performance by adjusting hyperparamters (such as $\lambda^{a}$) to be more biased towards audio. However, when reporting results, we take the audio performance from our best performing visual model, reporting a single model for audio-visual TIM.

\prepara\noindent\textbf{EPIC Details.} For EPIC-KITCHENS-100 and EPIC-SOUNDS, we include two extra CLS tokens for each visual query: $\texttt{[CLS]}_{verb, noun}^{v}$, along with classifiers $h^{v}_{\texttt{CLS}_{verb}}(\cdot)$ and $h^{v}_{\texttt{CLS}_{noun}}(\cdot)$.
We set the learning rate to 1e-4 and apply channel-wise dropout with $p=0.5$ directly to the raw input features, as well as to the encoded transformer input. 
We set $\lambda^{a}=0.01$ and $\lambda^{v}=1.0$. 
The low value of $\lambda^{a}$ is to alleviate early overfitting of the audio data, also observed in other works~\cite{xiao2020audiovisual}.

\prepara\noindent\textbf{AVE Details.} Due to the significantly smaller size of the AVE dataset, we reduce the model size for this dataset by halving the hidden dimension of all linear layers ($512$-D). We use an initial learning rate $5e-4$. We set all dropouts in the model to $p=0.1$. We set $N^{m}=10$ with $H_{f}=1.0$ to be consistent with other works. This results in a window size of $W=10$ seconds which is the full-length of the video in this dataset. We thus do not use any window stride $(H_{w})$ for this dataset. We apply AVGA~\cite{tian2018ave} to spatial visual features from VGG-19 before feeding it to the transformer. As this dataset does not contain distinct labels for audio and visual, we encourage the model to learn the single label set for both modalities by duplicating the query, i.e.\ using a $\texttt{[CLS]}$ for each modality, and combine their logits for training and inference.
We set $\lambda^{a}=1.0$ and $\lambda^{v}=1.0$.

\prepara\noindent\textbf{Perception Test Details.} We set the learning rate to 1e-4 and apply channel-wise dropout with $p=0.1$ to both the raw input features and encoded input sequence. 
We set $W=20$ seconds, $\lambda^{a}=1.0$, and $\lambda^{v}=1.0$.

\prepara\noindent\textbf{Detection Details.} Due to memory constraints, as opposed to using a single model to jointly train for all sub-tasks in recognition (visual and audio or verb, noun, action and audio in EPIC), we use a separate model for each individual sub-task, resulting in two different sets of model weights for detection and recognition. We also extend the number of layers in the transformer encoder from 4 to 6. The regression head consists of 2 layers with hidden dimension $D / 2$, followed by ReLU activations, and a final layer which outputs 2 numbers relating to the regressed boundaries, followed by a Sigmoid activation to scale the outputs between $[0, 1]$.

For Perception Sound and Action, we train for 100 epochs and use a $0.01$ confidence threshold and NMS $\sigma=0.1$. For EPIC, we train for 35 epochs and use a $0.03$ confidence threshold and NMS $\sigma=0.25$. All other hyper-parameters are consistent with the recognition models.

\end{document}